\pdfoutput=1
\documentclass{article}
\usepackage{amsmath}
\usepackage{amssymb}
\usepackage[mathscr]{euscript}
\usepackage{stmaryrd}
\usepackage[table,usenames,dvipsnames]{xcolor}
\definecolor{shadecolor}{gray}{0.9}

\usepackage[final,expansion=alltext]{microtype}
\usepackage[english]{babel}
\usepackage[parfill]{parskip}
\usepackage{afterpage}
\usepackage{framed}
\usepackage{verbatim}
\usepackage{setspace}
\usepackage{bbm}

{\endMakeFramed}

\usepackage{ragged2e}

\newcounter{parcount}

\usepackage[symbol]{footmisc}

\usepackage[inline]{enumitem}

\usepackage{graphicx}
\usepackage[labelfont={it}]{caption}
\usepackage[format=hang]{subcaption}
\usepackage{wrapfig}
\usepackage{caption}
\usepackage{subcaption}

\usepackage{chngcntr}

\usepackage{booktabs}
\usepackage{longtable}
\usepackage{hhline}
\usepackage{multirow}

\usepackage[algoruled]{algorithm2e}
\usepackage{listings}
\usepackage{fancyvrb}
\fvset{fontsize=\normalsize}

\usepackage{amsthm}
\newtheorem{proposition}{Proposition}

\newtheorem{theorem}{Theorem}

\usepackage[colorlinks,linktoc=all]{hyperref}
\usepackage[all]{hypcap}
\hypersetup{citecolor=MidnightBlue}
\hypersetup{linkcolor=black}
\hypersetup{urlcolor=MidnightBlue}

\usepackage[nameinlink]{cleveref}

\lstdefinestyle{mystyle}{
    commentstyle=\color{OliveGreen},
    keywordstyle=\color{BurntOrange},
    numberstyle=\tiny\color{black!60},
    stringstyle=\color{MidnightBlue},
    basicstyle=\ttfamily,
    breakatwhitespace=false,
    breaklines=true,
    captionpos=b,
    keepspaces=true,
    numbers=left,
    numbersep=5pt,
    showspaces=false,
    showstringspaces=false,
    showtabs=false,
    tabsize=2
}
\usepackage[colorinlistoftodos,
            prependcaption,
            textsize=small,
            backgroundcolor=yellow,
            linecolor=lightgray,
            bordercolor=lightgray]{todonotes}

\usepackage{soul}
\DeclareRobustCommand{\mb}[1]{\boldsymbol{#1}}

\DeclareMathOperator*{\argmax}{arg\,max}

\renewcommand{\mid}{~\vert~}

\newcommand{\mba}{\mathbf{a}}
\newcommand{\mbb}{\mathbf{b}}
\newcommand{\mbc}{\mathbf{c}}
\newcommand{\mbd}{\mathbf{d}}
\newcommand{\mbe}{\mathbf{e}}
\newcommand{\mbg}{\mathbf{g}}

\newcommand{\mbm}{\mathbf{m}}

\newcommand{\mbq}{\mathbf{q}}
\newcommand{\mbr}{\mathbf{r}}

\newcommand{\mbu}{\mathbf{u}}
\newcommand{\mbv}{\mathbf{v}}
\newcommand{\mbw}{\mathbf{w}}
\newcommand{\mbx}{\mathbf{x}}
\newcommand{\mby}{\mathbf{y}}
\newcommand{\mbz}{\mathbf{z}}

\newcommand{\mbA}{\mathbf{A}}

\newcommand{\mbC}{\mathbf{C}}

\newcommand{\mbE}{\mathbf{E}}

\newcommand{\mbG}{\mathbf{G}}
\newcommand{\mbH}{\mathbf{H}}
\newcommand{\mbI}{\mathbf{I}}
\newcommand{\mbJ}{\mathbf{J}}
\newcommand{\mbK}{\mathbf{K}}

\newcommand{\mbP}{\mathbf{P}}
\newcommand{\mbQ}{\mathbf{Q}}
\newcommand{\mbR}{\mathbf{R}}
\newcommand{\mbS}{\mathbf{S}}

\newcommand{\mbU}{\mathbf{U}}
\newcommand{\mbV}{\mathbf{V}}
\newcommand{\mbW}{\mathbf{W}}
\newcommand{\mbX}{\mathbf{X}}
\newcommand{\mbY}{\mathbf{Y}}

\newcommand{\mbpi}{\mb{\pi}}

\newcommand{\mbtheta}{\mb{\theta}}

\newcommand{\mbGamma}{\mb{\Gamma}}
\newcommand{\mbLambda}{\mb{\Lambda}}

\newcommand{\mbSigma}{\mb{\Sigma}}
\newcommand{\mbTheta}{\mb{\Theta}}

\newcommand{\bbE}{\mathbb{E}}

\newcommand{\cL}{\mathcal{L}}

\newcommand{\cN}{\mathcal{N}}
\newcommand{\cO}{\mathcal{O}}

\newcommand{\eA}{\mathscr{A}}

\newcommand{\eG}{\mathscr{G}}
\newcommand{\eH}{\mathscr{H}}

\newcommand{\beA}{\mb{\eA}}
\newcommand{\beG}{\mb{\eG}}
\newcommand{\beH}{\mb{\eH}}

\newcommand{\reals}{\mathbb{R}}

\newcommand{\method}{\textsc{SALT}\xspace}
\PassOptionsToPackage{compress, numbers}{natbib}
\usepackage{preamble/neurips_2023}
\usepackage[utf8]{inputenc}             % allow utf-8 input
\usepackage[T1]{fontenc}                % use 8-bit T1 fonts
\usepackage{url}                        % simple URL typesetting
\usepackage{booktabs}                   % professional-quality tables
\usepackage{amsfonts}                   % blackboard math symbols
\usepackage{nicefrac}                   % compact symbols for 1/2, etc.
\usepackage{microtype}                  % microtypography
\usepackage[dvipsnames]{xcolor}         % colors

\title{Switching Autoregressive Low-rank Tensor Models}

\author{Hyun Dong Lee \\
  Computer Science Department\\
  Stanford University\\
  \texttt{hdlee@stanford.edu} \\
  \and
  \textbf{Andrew Warrington} \\
  Department of Statistics\\
  Stanford University\\
  \texttt{awarring@stanford.edu} \\
  \and
  \textbf{Joshua I. Glaser} \\
  Department of Neurology\\
  Northwestern University\\
  \texttt{j-glaser@northwestern.edu} \\
  \and
  \textbf{Scott W. Linderman} \\
  Department of Statistics\\
  Stanford University\\
  \texttt{scott.linderman@stanford.edu}
}

\begin{document}

\maketitle

\begin{abstract}
An important problem in time-series analysis is modeling systems with time-varying dynamics.  Probabilistic models with joint continuous and discrete latent states offer interpretable, efficient, and experimentally useful descriptions of such data.  Commonly used models include autoregressive hidden Markov models (ARHMMs) and switching linear dynamical systems (SLDSs), each with its own advantages and disadvantages.  ARHMMs permit exact inference and easy parameter estimation, but are parameter intensive when modeling long dependencies, and hence are prone to overfitting.  In contrast, SLDSs can capture long-range dependencies in a parameter efficient way through Markovian latent dynamics, but present an intractable likelihood and a challenging parameter estimation task.  In this paper, we propose \emph{switching autoregressive low-rank tensor} (\method) models, which retain the advantages of both approaches while ameliorating the weaknesses.  \method parameterizes the tensor of an ARHMM with a low-rank factorization to control the number of parameters and allow longer range dependencies without overfitting.  We prove theoretical and discuss practical connections between \method, linear dynamical systems, and SLDSs.  We empirically demonstrate quantitative advantages of \method models on a range of simulated and real prediction tasks, including behavioral and neural datasets.  Furthermore, the learned low-rank tensor provides novel insights into temporal dependencies within each discrete state.
\end{abstract}

\section{Introduction}
Many time series analysis problems involve jointly segmenting data and modeling the time-evolution of the system within each segment.  For example, a common task in computational ethology~\citep{datta2019call} --- the study of natural behavior --- is segmenting videos of freely moving animals into states that represent distinct behaviors, while also quantifying the differences in dynamics between states~\citep{wiltschko2015mapping, costacurta2022distinguishing}.  Similarly, discrete shifts in the dynamics of neural activity may reflect changes in underlying brain state~\citep{saravani2019dynamic, recanatesi2022metastable}.  Model-based segmentations are experimentally valuable, providing an unsupervised grouping of neural or behavioral states together with a model of the dynamics within each state.

One common probabilistic state space model for such analyses is the \emph{autoregressive hidden Markov model} (ARHMM)~\citep{ephraim1989application}. For example, MoSeq~\citep{wiltschko2015mapping} uses ARHMMs for unsupervised behavioral analysis of freely moving animals.  ARHMMs learn a set of linear autoregressive models, indexed by a discrete state, to predict the next observation as a function of previous observations.  Inference in ARHMMs then reduces to inferring which AR process best explains the observed data at each timestep (in turn also providing the segmentation).  The simplicity of ARHMMs allows for exact state inference via message passing, and closed-form updates for parameter estimation using expectation-maximization (EM).  However, the ARHMM requires high order autoregressive dependencies to model long timescale dependencies, and its parameter complexity is quadratic in the data dimension, making it prone to overfitting.

\emph{Switching linear dynamical systems} (SLDS)~\citep{ghahramani2000variational} ameliorate some of the drawbacks of the ARHMM by introducing a low-dimensional, continuous latent state.  These models have been used widely throughout neuroscience~\citep{saravani2019dynamic, petreska2011dynamical, linderman2019hierarchical, glaser2020recurrent, nair2023approximate}.  Unlike the ARHMM, the SLDS can capture long timescale dependencies through the dynamics of the continuous latent state, while also being much more parameter efficient than ARHMMs.  However, exact inference in SLDSs is intractable due to the exponential number of potential discrete state paths governing the time-evolution of the continuous latent variable.  This intractability has led to many elaborate and specialized approximate inference techniques~\citep{ghahramani2000variational, barber2006expectation, fox2009bayesian, murphy2001rao, linderman2017bayesian, zoltowski2020}.  Thus, the SLDS gains parameter efficiency at the expense of the computational tractability and statistical simplicity of the ARHMM.

\begin{figure*}[!t]
\begin{center}
	\includegraphics[width=\textwidth]{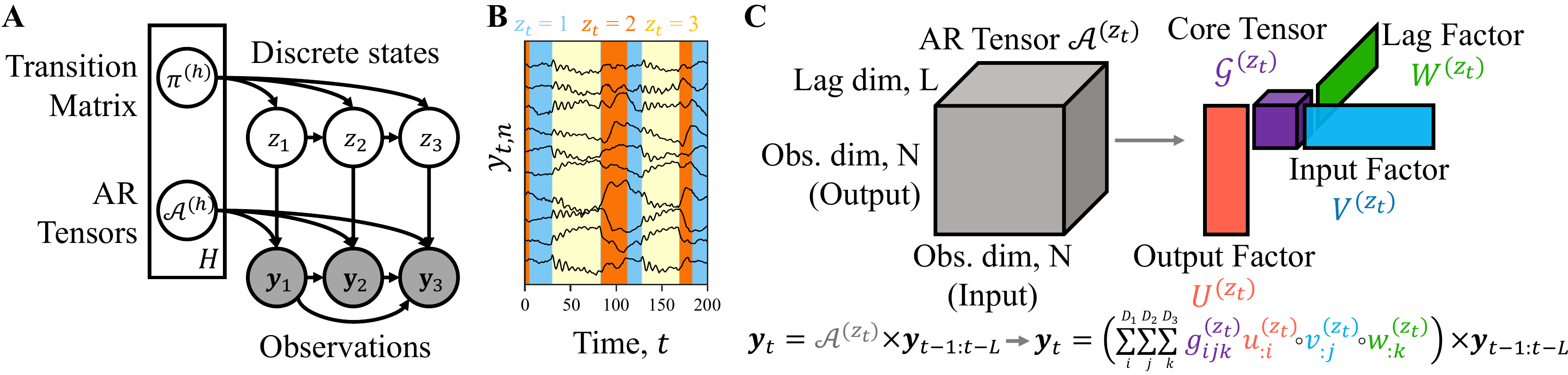}
    \caption{\textbf{\method imposes a low-rank constraint on the autoregressive tensor}: \textbf{(A)} The probabilistic graphical model of an ARHMM.  \textbf{(B)} An example multi-dimensional time series generated from an ARHMM.  Background color indicates which discrete state (and hence autoregressive tensor) was selected at each time. \textbf{(C)} In \method, each autoregressive dynamics tensor of an ARHMM is parameterized as a low-rank tensor.}
    \label{fig:SALT}
\end{center}
\end{figure*}

We propose a new class of unsupervised probabilistic models that we call \emph{switching autoregressive low-rank tensor} (SALT) models.  
Our key insight is that when you marginalize over the latent states of a linear dynamical system, you obtain an autoregressive model with full history dependence. 
However, the autoregressive dependencies are not arbitrarily complex --- they factor into a low-rank tensor that can be well-approximated with a finite-history model (Proposition~\ref{prop:lds}). 
\method models are constrained ARHMMs that leverage this insight. Rather than allowing for arbitrary autoregressive dependencies, \method models are constrained to be low-rank.  
Thus, \method models inherit the parsimonious parameter complexity of SLDS as well as the ease of inference and estimation of ARHMMs.
We demonstrate the advantages of \method models empirically using synthetic data as well as real neural and behavioral time series. 
Finally, in addition to improving predictive performance, we show how the low-rank nature of SALT models can offer new insights into complex systems, like biological neural networks. Source code is available at \url{https://github.com/lindermanlab/salt}.

\section{Background}
This section introduces the notation used throughout the paper and describes preliminaries on low-rank tensor decomposition, vector autoregressive models, switching autoregressive models, linear dynamical systems, and switching linear dynamical systems.

\paragraph{Notation}
We follow the notation of \citet{kolda2009tensor}.
We use lowercase letters for scalar variables (e.g. $a$), uppercase letters for scalar constants (e.g.~$A$), boldface lowercase letters for vectors (e.g. $\mba$), boldface uppercase letters for matrices (e.g. $\mbA$), and boldface Euler script for tensors of order three or higher (e.g. $\beA$).
We will use the shorthand~$\mba_{1:T} \in \reals^{N \times T}$ to denote a time series~$\mba_1 \in \reals^{N}, \ldots, \mba_T \in \reals^{N}$. 
We use $\mbA_{i::}$, $\mbA_{:j:}$, and $\mbA_{::k}$ to denote the horizontal, lateral, and frontal slices respectively of a three-way tensor $\beA$.
Similarly, we use $\mba_{i:}$ and $\mba_{:j}$ to denote the $i^{th}$ row and $j^{th}$ column of a matrix $\mbA$.
$\mba \circ \mbb$ represents the vector outer product between vectors $\mba$ and $\mbb$.
The $n$-mode tensor-matrix (tensor-vector) product is represented as $\beA \times_n \mbA$ ($\beA \bar{\times}_n \mba$).
We represent the mode-$n$ matricization of a tensor $\beG$ as $\beG_{(n)}$.
We define $\times_{j,k}$ to be a tensor-matrix product over the $j^{th}$ and $k^{th}$ slices of the tensor.
For example, given a three-way tensor $\beA \in \reals^{D_1 \times D_2 \times D_3}$ and a matrix $\mbX \in \reals^{D_2 \times D_3}$, $\beA \times_{2,3} \mbX = \sum_{j=1}^{D_2} \sum_{k=1}^{D_3} \mba_{:jk} x_{jk}$.

\paragraph{Tensor Decomposition}
For $\beA \in \reals^{N_1 \times N_2 \times N_3}$, the Tucker decomposition is defined as,
\begin{align}
    \beA &= \sum_{i=1}^{D_1} \sum_{j=1}^{D_2} \sum_{k=1}^{D_3} g_{ijk} \, \mbu_{:i} \circ \mbv_{:j} \circ \mbw_{:k}, \label{equ:background:tucker}
\end{align}
where $\mbu_{:i}$, $\mbv_{:j}$, and $\mbw_{:k}$ are the columns of the factor matrices~$\mbU \in \reals^{N_1 \times D_1}$,~$\mbV \in \reals^{N_2 \times D_2}$, and~$\mbW \in \reals^{N_3 \times D_3}$, respectively, and $g_{ijk}$ are the entries in the core tensor~$\beG \in \reals^{D_1 \times D_2 \times D_3}$.

The CANDECOMP/PARAFAC (CP) decomposition is a special case of the Tucker decomposition, with $D_1=D_2=D_3$ and a diagonal core tensor $\beG$.

\paragraph{Vector autoregressive models}
Let~$\mby_{1:T}$ denote a multivariate time series with~$\mby_t \in \reals^N$ for all~$t$. An order-$L$ vector autoregressive (VAR) model with Gaussian innovations is defined by, 
\begin{align}
%    \mby_{t} &\sim \cN \left(\sum_{l=1}^L \mbA^{\!(l)} \mby_{t-l} + \mbb, \, \mbQ \right),
    \mby_{t} &\sim \cN \left(\beA \times_{2,3} \mby_{t-1:t-L} + \mbb, \, \mbR \right),
\end{align}
where~${\beA \in \reals^{N \times N \times L}}$ is the autoregressive tensor, whose frontal slice $\mbA_{::l}$ is the dynamics matrix for lag~$l$, ~$\mbb \in \reals^{N}$ is the bias, and ~$\mbR \in \reals^{N \times N}_{\succeq 0}$ is a positive semi-definite covariance matrix.
The parameters~$\mbTheta = (\beA, \mbb, \mbR)$ are commonly estimated via ordinary least squares~\citep{hamilton2020time}.
%where~${\mbA^{\!(l)} \in \reals^{N \times N}}$ is the dynamics matrix for lag~$l$,~$\mbb \in \reals^{N}$ is the bias, and~$\mbQ \in \reals^{N \times N}_{\succeq 0}$ is a positive definite covariance matrix.
%Typically, the parameters~$\mbTheta = (\{\mbA^{(l)}\}_{l=1}^L, \mbb, \mbQ)$ are estimated via ordinary least squares~\citep{hamilton1994}.

\paragraph{Switching autoregressive models} 
One limitation of VAR models is that they assume the time series is stationary; i.e. that one set of parameters holds for all time steps. Time-varying autoregressive models allow the autoregressive process to change at various time points. One such VAR model, referred to as a switching autoregressive model or autoregressive hidden Markov model (ARHMM), switches the parameters over time according to a discrete latent state~\citep{ephraim1989application}.
Let~$z_t \in \{1,\ldots,H\}$ denote the discrete state at time~$t$, an ARHMM defines the following generative model,
\begin{align}
    z_t \sim \mathrm{Cat}(\mbpi^{(z_{t-1})}), \qquad  \mby_t \sim \cN \left(\beA^{(z_t)} \times_{2,3} \mby_{t-1:t-L} + \mbb^{(z_t)}, \, \mbR^{(z_t)} \right), \label{eq:z}
\end{align}
where~$\mbpi^{(h)} \in \{\mb\pi^{(h)}\}_{h=1}^H$ is the the~$h$-th row of the discrete state transition matrix.

A switching VAR model is simply a type of hidden Markov model, and as such it is easily fit via the expectation-maximization (EM) algorithm within the Baum-Welch algorithm. 
The M-step amounts to solving a weighted least squares problem.

\paragraph{Linear dynamical systems} 
The number of parameters in a VAR model grows as~$\cO(N^2L)$.
For high-dimensional time series, this can quickly become intractable. 
Linear dynamical systems (LDS)~\citep{murphy2012machine} offer an alternative means of modeling time series via a continuous latent state~$\mbx_t \in \reals^D$,
\begin{align}
    \mbx_t \sim \cN(\mbA \mbx_{t-1} + \mbb, \, \mbQ),  \qquad \mby_t \sim \cN(\mbC \mbx_t + \mbd, \, \mbR),  \label{eq:lds}
\end{align}
where $\mbQ \in \reals^{D \times D}_{\succeq 0}$ and $\mbR \in \reals^{N \times N}_{\succeq 0}$.
Here, the latent states follow a first-order VAR model, and the observations are conditionally independent given the latent states. 
As we discuss in Section \ref{sec:methods:slds}, marginalizing over the continuous latent states renders~$\mby_t$ dependent on the preceding observations, just like in a high order VAR model.

Compared to the VAR model, however, the LDS has only~$\cO(D^2 + ND + N^2)$ parameters if~$\mbR$ is a full covariance matrix.  This further reduces to $\cO(D^2 + ND)$ if $\mbR$ is diagonal.  As a result, when~$D \ll N$, the LDS has many fewer parameters than a VAR model.  Thanks to the linear and Gaussian assumptions of the model, the parameters can be easily estimated via EM, using the Kalman smoother to compute the expected values of the latent states.

\paragraph{Switching linear dynamical systems} A switching LDS combines the advantages of the low-dimensional continuous latent states of an LDS, with the advantages of discrete switching from an ARHMM. Let~$z_t \in \{1,\ldots, H\}$ be a discrete latent state with Markovian dynamics~\eqref{eq:z}, and let it determine some or all of the parameters of the LDS (e.g.~$\mbA$ would become~$\mbA^{\!(z_t)}$ in~\eqref{eq:lds}).  We note that SLDSs often use a \emph{single-subspace}, where $\mbC$, $\mbd$ and $\mbR$ are shared across states, reducing parameter complexity and simplifying the optimization. 

Unfortunately, parameter estimation is considerably harder in SLDS models. The posterior distribution over all latent states,~$p(\mbz_{1:T}, \mbx_{1:T} \mid \mby_{1:T}, \mbTheta)$, where $\mbTheta$ denotes the parameters, is intractable \citep{lerner2003hybrid}. Instead, these models are fit via approximate inference methods like MCMC~\citep{fox2009bayesian,linderman2017bayesian}, variational EM~\citep{ghahramani2000variational, zoltowski2020}, particle EM~\citep{murphy2001rao, doucet2001particle}, or other approximations \citep{barber2006expectation}.  We look to define a model that enjoys the benefits of SLDSs, but avoids the inference and estimation difficulties.
\section{SALT: Switching Autoregressive Low-rank Tensor Models}
\label{sec:methods}

Here we formally introduce \method models.  We begin by defining the generative model (also illustrated in Figure \ref{fig:SALT}), and describing how inference and model fitting are performed.  We conclude by drawing connections between \method and SLDS models.

\subsection{Generative Model}

\method factorizes each autoregressive tensor $\beA^{(h)}$ for $h\in\lbrace 1, \ldots, H \rbrace$ of an ARHMM as a product of low-rank factors.  Given the current discrete state $z_t$, each observation $\mby_t \in \reals^N$ is modeled as being normally distributed conditioned on $L$ previous observations $\mby_{t-1:t-L}$,
\begin{align}
    z_t &\sim \mathrm{Cat}\left(\mbpi^{(z_{t-1})}\right), \label{equ:salt:transition} \\
    \mby_t &\overset{\text{i.i.d.}}{\sim} \mathcal{N}\left(\beA_{\method}^{(z_t)} \times_{2,3} \mby_{t-1:t-L} + \mbb^{(z_t)} ,\pmb{\Sigma}^{(z_t)} \right), \label{equ:salt:likelihood}\\
    \beA_{\method}^{(z_t)} &= \sum_{i=1}^{D_1} \sum_{j=1}^{D_2} \sum_{k=1}^{D_3} g_{ijk}^{(z_t)} \, \mbu_{:i}^{(z_t)} \circ \mbv_{:j}^{(z_t)} \circ \mbw_{:k}^{(z_t)}, \label{equ:salt:A_def}
\end{align}
where $\mbu_{:i}^{(z_t)}$, $\mbv_{:j}^{(z_t)}$, and $\mbw_{:k}^{(z_t)}$ are the columns of the factor matrices ${\mbU^{(z_t)} \in \reals^{N \times D_1}}$, ${\mbV^{(z_t)} \in \reals^{N \times D_2}}$, and~${\mbW^{(z_t)} \in \reals^{L \times D_3}}$, respectively, and $g_{ijk}^{(z_t)}$ are the entries in the core tensor~${\beG^{(z_t)} \in \reals^{D_1 \times D_2 \times D_3}}$. The vector $\mb b^{(z_t)} \in \reals^{N}$ and positive definite matrix $\pmb{\Sigma}^{(z_t)} \in \reals^{N \times N}_{\succeq 0}$ are the bias and covariance for state $z_t$.  Without further restriction this decomposition is a Tucker decomposition~\citep{kolda2009tensor}.  If $D_1 = D_2 = D_3$ and $\beG_{z_t}$ is diagonal, it corresponds to a CP decomposition~\citep{kolda2009tensor}.  We refer to ARHMM models with these factorizations as Tucker-\method and CP-\method respectively.  Note that herein we will only consider models where $D_1 = D_2 = D_3 = D$, where we refer to $D$ as the ``rank'' of the \method model (for both Tucker-\method and CP-\method).  In practice, we find that models constrained in this way perform well, and so this constraint is imposed simply to reduce the search space of models.  This constraint can also be easily relaxed.  

Table \ref{tab:param} shows the number of parameters for order-$L$ ARHMMs, SLDSs, and \method. Focusing on the lag dependence, the number of ARHMM parameters grows as $\cO(HN^2L)$, whereas \method grows as only $\cO(HDL)$ with $D \ll N$.  \method can also make a simplifying single-subspace constraint, where certain emission parameters are shared across discrete states.  

\paragraph{Low-dimensional Representation}  Note that \method implicitly defines a low-dimensional continuous representation, analogous to the continuous latent variable in SLDS,
\begin{equation}
    \mbx_t = \left(\sum_{j=1}^{D_2} \sum_{k=1}^{D_3} \mbg_{:jk}^{(z_t)} \circ \mbv_{:j}^{(z_t)} \circ \mbw_{:k}^{(z_t)}\right) \times_{2,3} \mby_{t-1:t-L}.
\end{equation}
The vector $\mbx_t \in \mathbb{R}^{D_1}$ is multiplied by the output factors, $\mbU^{(z_t)}$, to obtain the mean of the next observation.  These low-dimensional vectors can be visualized as in SLDS and used to further interrogate the learned dynamics, as we show in Figure \ref{fig:slds_simulated}.

\subsection{Model Fitting and Inference}
Since \method models are ARHMMs, we can apply the expectation-maximization (EM) algorithm to fit model parameters and perform state space inference.  We direct the reader to \citet{murphy2012machine} for a detailed exposition of EM and include only the key points here.

The E-step solves for the distribution over latent variables given observed data and model parameters.  For \method, this is the distribution over $z_t$, denoted ${\omega_{t}^{(h)} = \bbE[z_t = h \mid \mby_{1:T}, \mbtheta]}$.  This can be computed exactly with the forward-backward algorithm, which is fast and stable.  The marginal likelihood can be evaluated exactly by taking the product across $t$ of expectations of \eqref{equ:salt:likelihood} under $\omega_{t}^{(h)}$.

The M-step then updates the parameters of the model given the distribution over latent states.  For \method, the emission parameters are $\mbtheta = \lbrace \mbU^{(h)}, \mbV^{(h)}, \mbW^{(h)}, \beG^{(h)}, \mb b^{(h)}, \pmb{\Sigma}^{(h)}, \mbpi^{(h)} \rbrace_{h=1}^H$.  We use closed-form coordinate-wise updates to maximize the expected log likelihood evaluated in the E-step.  Each factor update amounts to solving a weighted least squares problem. We include just one update step here for brevity, and provide all updates in full in Appendix \ref{sec:update}.  Assuming here that $\mbb^{(h)} = \textbf{0}$ for simplicity, the update rule for the lag factors is as follows:
\begin{align}
    \mbw^{(h) \star} &= \left(\sum_t \omega_{t}^{(h)} \widetilde{\mbX}_{t}^{(h) \top} (\mb\Sigma^{(h)})^{-1} \widetilde{\mbX}_{t}^{(h)} \right)^{-1} \left(\sum_t \omega_{t}^{(h)} \widetilde{\mbX}_{t}^{(h) \top} (\mb\Sigma^{(h)})^{-1} \mby_t \right)
\end{align}
where $\widetilde{\mbX}_{t}^{(h)} = \mbU^{(h)} \beG_{(1)}^{(h)} (\mbV^{(h) \top} \mby_{t-1:t-L} \otimes \mbI_{D_3})$ and $\mbw^{(h) \star} = \mathrm{vec}(\mbW^{(h)})$.  Crucially, these coordinate wise updates are exact, and so we recover the fast and monotonic convergence of EM.  

\newcolumntype{C}[1]{>{\centering\let\newline\\\arraybackslash\hspace{0pt}}m{#1}}

\begin{table}[t!]
\centering
\begin{tabular}{llr}
\toprule
\textbf{Model}                  & \textbf{Parameter Complexity}             & \textbf{(Example from Section \ref{sec:results:ce})}                      \\     \midrule
SLDS                            & $\cO(ND+HD^2)$                            & 2.8K \qquad\quad\quad\quad                                                             \\
CP-\method                      & $\cO(H(ND+LD))$                           & 8.1K \qquad\quad\quad\quad                                                             \\      
Tucker-\method                  & $\cO(H(ND+LD+D^3))$                       & 17.4K \qquad\quad\quad\quad                                                            \\
Order-$L$ ARHMM                 & $\cO(HN^2L)$                              & 145.2K \qquad\quad\quad\quad                                                           \\
\bottomrule
\end{tabular}
\vspace*{0.2cm}
\caption{Comparison of number of parameters for the methods we consider.  We exclude covariance matrix parameters, as the parameterization of the covariance matrix is independent of method.}
\label{tab:param}
\vspace{-0.25cm}
\end{table}

\subsection{Connections Between \method and Switching Linear Dynamical Systems}
\label{sec:methods:slds}
\method is not only an intuitive regularization for ARHMMs, it is grounded in a mathematical correspondence between autoregressive models and linear dynamical systems. 

\begin{proposition}[Low-Rank Tensor Autoregressions Approximate Stable Linear Dynamical Systems]
    \label{prop:lds}
    % \ref{prop:lds}
    Consider a stable linear time-invariant Gaussian dynamical system.  We define the steady-state Kalman gain matrix as $\mbK = \lim_{t \rightarrow \infty} \mbK_t$, and $\mb\Gamma = \mbA(\mbI - \mbK\mbC)$.  The matrix $\mb\Gamma \in \reals^{D \times D}$ has eigenvalues $\lambda_1, \ldots, \lambda_D$. Let $\lambda_{\mathsf{max}} = \max_d |\lambda_d|$; for a stable LDS, $\lambda_{\mathsf{max}} < 1$~\emph{\citep{davis1985control}}. Let $n$ denote the number of real eigenvalues and $m$ the number of complex conjugate pairs. Let $\hat{\mby}_t^{(\mathsf{LDS})} = \mathbb{E}[\mby_t \mid \mby_{<t}]$ denote the predictive mean under a steady-state LDS, and $\hat{\mby}_t^{(\mathsf{SALT})}$ the predictive mean under a SALT model. An order-$L$ Tucker-SALT model with rank $n + 2m$, or a CP-SALT model with rank $n + 3m$, can approximate the predictive mean of the steady-state LDS with error $\|\hat{\mby}_t^{(\mathsf{LDS})} - \hat{\mby}_t^{(\mathsf{SALT})}\|_\infty = \cO(\lambda_{\mathsf{max}}^L)$.
\end{proposition}
\begin{proof}
We give a sketch of the proof here and a full proof in Appendix \ref{sec:lds_tr}.  The analytic form of $\bbE \left[ \mby_t \mid \mby_{<t} \right]$ is a linear function of $\mby_{t-l}$ for $l=1,\ldots,\infty$.  For this proof sketch, consider the special case where $\mbb = \mbd = \boldsymbol{0}$. Then the coefficients of the linear function are $\mbC \mbGamma^l \mbK$.  As all eigenvalues of~$\mb\Gamma$ have magnitude less than one, the coefficients decay exponentially in $l$.  We can therefore upper bound the approximation error introduced by truncating the linear function to $L$ terms to $\mathcal{O}(\lambda_{\mathrm{max}}^L)$.  To complete the proof, we show that the truncated linear function can be represented exactly by a tensor regression with at most a specific rank.  Thus, only truncated terms contribute to the error.  
\end{proof}

This proposition shows that the steady-state predictive distribution of a stable LDS can be approximated by a low-rank tensor autoregression, with a rank determined by the eigenspectrum of the LDS.  We validate this proposition experimentally in Section \ref{sec:results:lds}.  Note as well that the predictive distribution will converge to a fixed covariance, and hence can also be exactly represented by the covariance matrices~$\mbSigma^{(h)}$ estimated in \method models.  

With this foundation, it is natural to hypothesize that a \textit{switching} low-rank tensor autoregression like \method could approximate a \textit{switching} LDS. There are two ways this intuition could fail: first, if the dynamics in a discrete state of an SLDS are unstable, then~\Cref{prop:lds} would not hold; second, after a discrete state transition in an SLDS, it may take some time before the dynamics reach stationarity. We empirically test how well \method approximates an SLDS in Section~\ref{sec:results} and find that, across a variety of datasets, \method obtains commensurate performance with considerably simpler inference and estimation algorithms.
\section{Related Work}

\paragraph{Low-rank tensor decompositions of time-invariant autoregressive models}
Similar to this work, \citet{wang2021high} also modeled the transition matrices as a third-order tensor~$\beA \in \reals^{N \times N \times L}$ where the~$\mbA_{::l}$ is the~$l$-th dynamics matrix.  They then constrained the tensor to be low-rank via a Tucker decomposition, as defined in \eqref{equ:background:tucker}.  However, unlike \method, their model was time-invariant, and they did not have an ARHMM structure or make connections to the LDS and SLDS, as in Proposition~\ref{prop:lds}.

\paragraph{Low-rank tensor decompositions of time-varying autoregressive models} Low-rank tensor-based approaches have also been used to model time-varying AR processes~\citep{tvart, zhang2021bayesian}.  \citet{tvart} introduced TVART, which first splits the data into $T$ contiguous fixed-length segments, each with its own AR-1 process.  TVART can be thought of as defining a $T \times N \times N$ ARHMM dynamics tensor and progressing through discrete states at fixed time points.  This tensor is parameterized using the CP decomposition and optimized using an alternating least squares algorithm, with additional penalties such that the dynamics of adjacent windows are similar.  By contrast, SALT automatically segments, rather than windows, the time-series into learned and re-usable discrete states.  

\citet{zhang2021bayesian} constructed a Bayesian model of higher-order AR matrices that can vary over time.  First, $H$ VAR dynamics tensors are specified, parameterized as third-order tensors with a rank-1 CP decomposition.  The dynamics at any given time are then defined as a weighted sum of the tensors, where the weights have a prior density specified by an Ising model.  Finally, inference over the weights is performed using MCMC.  This method can be interpreted as a factorial ARHMM and hence offers substantial modeling flexibility, but it sacrifices computational tractability when $H$ is large.  

\paragraph{Low-rank tensor decompositions of neural networks}
Low-rank tensor decomposition methods have also been used to make neural networks more parameter efficient.
\citet{novikov2015tensorizing} used the tensor-train decomposition~\citep{oseledets2011tensor} on the dense weight matrices of the fully-connected layers to reduce the number of parameters.
\citet{yu2017long} and \citet{qiu2021memory} applied the tensor-train decomposition to the weight tensors for polynomial interactions between the hidden states of recurrent neural networks~(RNNs) to efficiently capture high-order temporal dependencies.
Unlike switching models with linear dynamics, recurrent neural networks have dynamics that are hard to interpret, their state estimates are not probabilistic, and they do not provide experimentally useful data segmentations.

\paragraph{Linear dynamical systems and low-rank linear recurrent neural networks}
\citet{valente2022probing} recently examined the relationship between LDSs and low-rank linear RNNs.
They provide the conditions under which low-rank linear RNNs can exactly model the first-order autoregressive distributions of LDSs, and derive the transformation to convert between model classes under those conditions.  This result has close parallels to Proposition \ref{prop:lds}.  Under the conditions identified by \citet{valente2022probing}, the approximation in Proposition \ref{prop:lds} becomes exact with just one lag term.  However, when those conditions are not satisfied, we show that one still recovers an LDS approximation with a bounded error that decays exponentially in the number of lag terms.
\section{Results}
\label{sec:results}

We now empirically validate \method by first validating the theoretical claims made in Section \ref{sec:methods}, and then apply \method to two synthetic examples to compare \method to existing methods.  We conclude by using \method to analyze real mouse behavioral recordings and \emph{C. elegans} neural recordings.

\subsection{\method Faithfully Approximates LDS}
\label{sec:results:lds}
\begin{figure*}[!t]
\begin{center}
	\includegraphics[width=\textwidth]{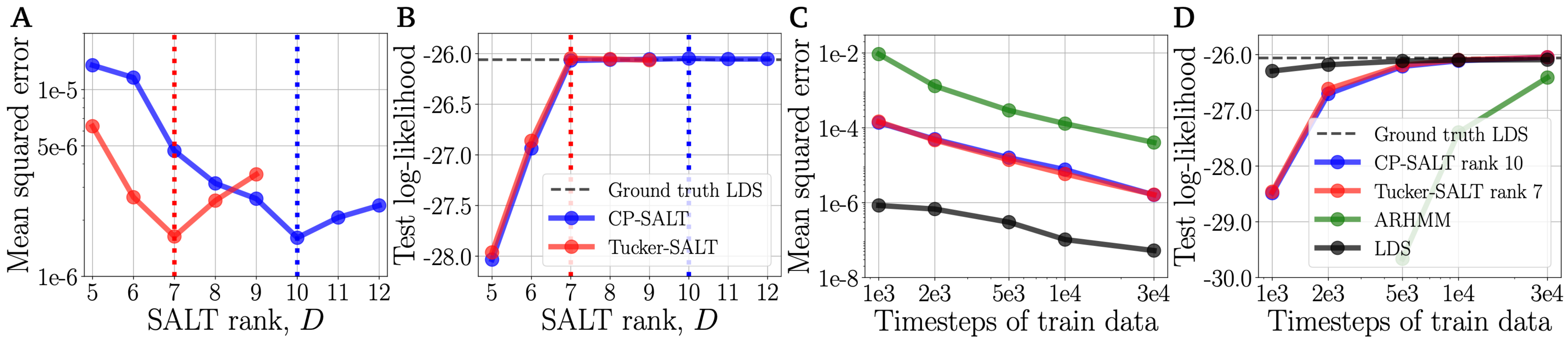}
    \vspace*{-0.5cm}
    \caption{\textbf{\method approximates LDS}: Data simulated from an LDS for which $n=1$ and $m=3$ (see Proposition \ref{prop:lds}).  \textbf{(A-B)}:  Average mean squared error of the autoregressive tensor corresponding to the LDS simulation and the log-likelihood of test data, as a function of \method rank.  According to Proposition \ref{prop:lds}, to model the LDS Tucker-\method and CP-\method require 7 and 10 ranks respectively (indicated by vertical dashed lines). \textbf{(C-D)}: Mean squared error of the learned autoregressive tensor and log-likelihood of test data as a function of training data.  }
    \label{fig:lds_simulated}
\end{center}
\end{figure*}

To test the theoretical result that \method can closely approximate a linear dynamical system, we fit \method models to data sampled from an LDS.  The LDS has $D=7$ dimensional latent states with random rotational dynamics, where $\mb\Gamma$ has $n=1$ real eigenvalue and $m=3$ pairs of complex eigenvalues, and $N=20$ observations with a random emission matrix.  %We sample $T=$30,000 timesteps from the LDS. 

For Figure~\ref{fig:lds_simulated}, we trained CP-\method and Tucker-\method with $L=50$ lags and varying ranks.  We first analyzed how well \method reconstructed the parameters of the autoregressive dynamics tensor.  As predicted by Proposition \ref{prop:lds}, Figure \ref{fig:lds_simulated}A shows that the mean squared errors between the \method tensor and the autoregressive tensor corresponding to the simulated LDS are the lowest when the ranks of CP-\method and Tucker-\method are $n+3m=10$ and $n+2m=7$ respectively.  We then computed log-likelihoods on 5,000 timesteps of held-out test data (Figure \ref{fig:lds_simulated}B).  Interestingly, the predictive performance of both CP-\method and Tucker-\method reach the likelihood of the ground truth LDS model with rank $n+2m=7$, suggesting that sometimes smaller tensors than suggested by Proposition \ref{prop:lds} may still be able to provide good approximations to the data.  We also show in Figures \ref{fig:lds_simulated}C and  \ref{fig:lds_simulated}D that, as predicted, \method models require much less data to fit than ARHMMs. We show extended empirical results and discussion on \Cref{prop:lds} in \Cref{subsec:extended_theory_experiments}.

\begin{figure}[!t]
\begin{center}
	\includegraphics[width=\textwidth]{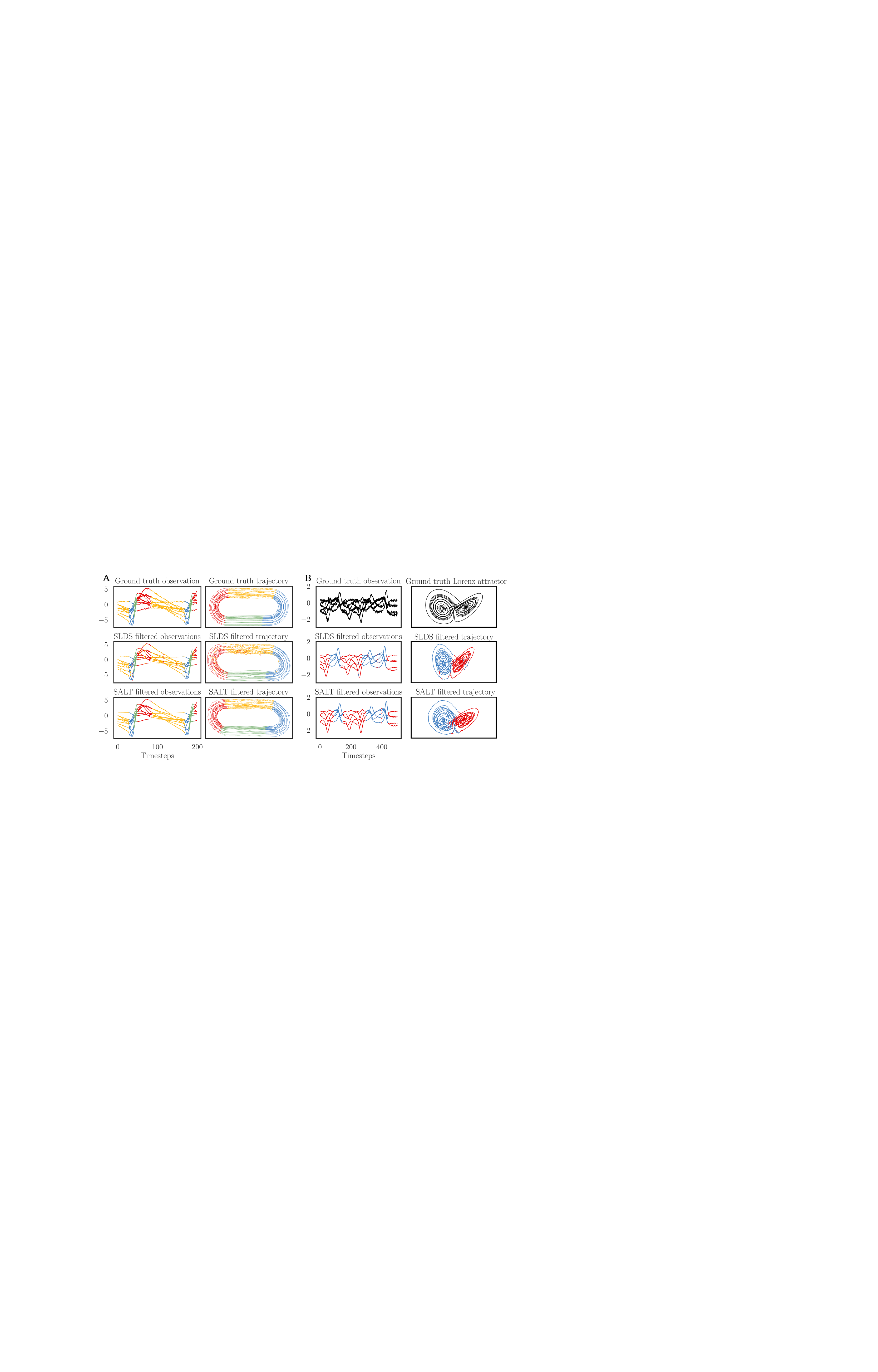}
    \vspace*{-0.5cm}
    \caption{\textbf{\method reconstructs simulated SLDS data and Lorenz attractor}: \textbf{(A)} Ground truth low-dimensional trajectory generated from a recurrent SLDS, as in \cite{linderman2017bayesian}, and 10-dimensional observations are generated from these latents. \textbf{(B)} The ground truth low-dimensional trajectory is generated from a Lorenz attractor, and 20-dimensional observations are generated from thse latents. Only 5 dimensions are shown for visual clarity. \textbf{(Top)}: Ground truth observations and trajectories. \textbf{(Middle and bottom)}: Fit trajectories and filtered observations from SLDS and \method models. Colors indicate discrete state for ground truth (in \textbf{(A)}) and fitted models. SLDS and \method find comparable filtered trajectories and observations.
    Note: we manually align latent trajectories for ease of inspection as both SLDS and \method are only identifiable in the latent space up to a linear transformation. }
    \label{fig:slds_simulated}
\end{center}
\vspace*{-0.3cm}
\end{figure}

\subsection{Synthetic Switching LDS Examples}
\label{sec:results:syn}

Proposition \ref{prop:lds} quantifies the convergence properties of low-rank tensor regressions when approximating stable LDSs.  Next we tested how well \method can approximate the more expressive \emph{switching} LDSs.  We first applied \method to data generated from a recurrent SLDS~\citep{linderman2017bayesian}, where the two-dimensional ground truth latent trajectory resembles a NASCAR\textsuperscript{\textregistered} track (Figure \ref{fig:slds_simulated}A). \method accurately reconstructed the ground truth filtered trajectories and discrete state segmentation, and yielded very similar results to an SLDS model.  We also tested the ability of \method to model nonlinear dynamics -- specifically, a Lorenz attractor -- which SLDSs are capable of modeling. Again, \method accurately reconstructed ground truth latents and observations, and closely matched SLDS segmentations. These results suggest that \method models provide a good alternative to SLDS models. Finally, in \Cref{subsec:tvart_comparison}, we used SLDS-generated data to compare \method and TVART~\citep{tvart}, another tensor-based method for modeling autoregressive processes, and find that \method more accurately reconstructed autoregressive dynamics tensors than TVART.
%find that \method outperforms TVART according to a number of metrics. 

\subsection{Modeling Mouse Behavior}
\label{sec:results:mouse}

\begin{figure}[!t]
\begin{center}
	\includegraphics[width=\textwidth]{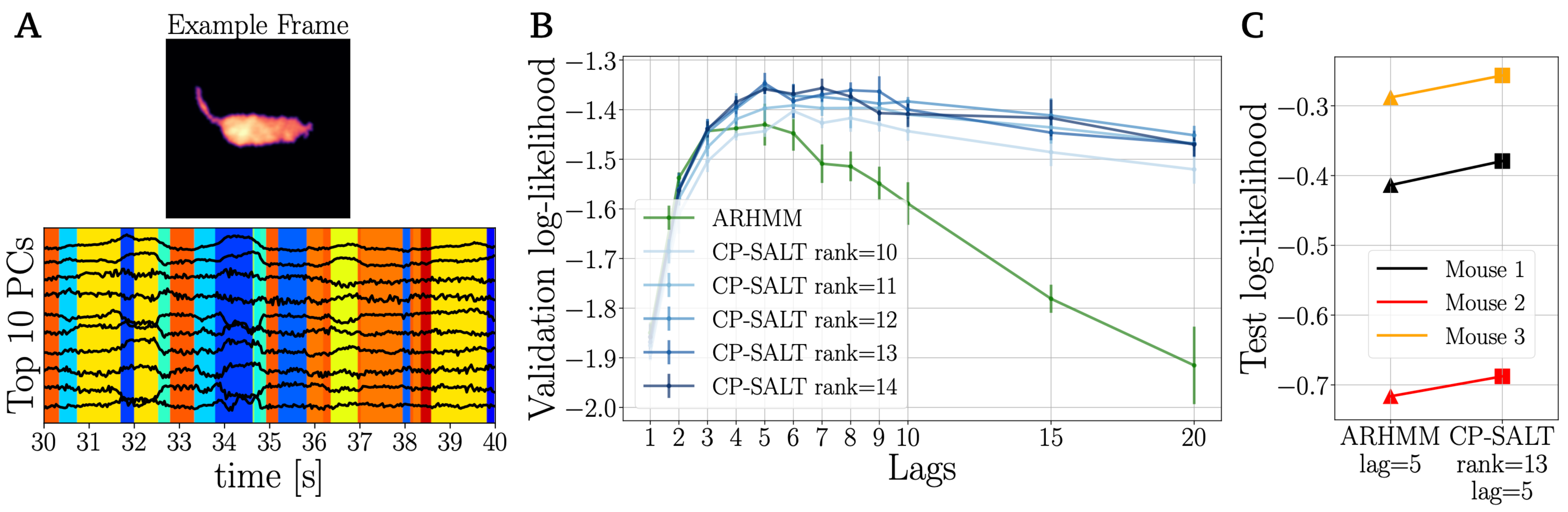}
    \vspace*{-0.2cm}
    \caption{\textbf{CP-\method consistently outperforms ARHMM on mouse behavior videos and segments data into distinct behavioral syllables}: \textbf{(A)} An example frame from the MoSeq dataset. The models were trained on the top 10 principal components of the video frames from three mice. \textbf{(B)} CP-\method and ARHMM trained with different ranks and lags. Mean and standard deviation across five seeds evaluated on a validation set are shown. CP-\method parameterization prevents overfitting for larger lags. \textbf{(C)} Test log-likelihood, averaged across 5 model fits, computed from the best ARHMM and CP-\method hyperparameters in (B).  CP-\method outperforms ARHMM across all three mice.}
    \label{fig:moseq_result}
\end{center}
\vspace*{-0.2cm}
\end{figure}

Next we considered a video segmentation problem commonly faced in the field of computational neuroethology~\citep{datta2019call}.
\citet{wiltschko2015mapping} collected videos of mice freely behaving in a circular open field.
They projected the video data onto the top 10 principal components (Figure \ref{fig:moseq_result}A) and used an ARHMM to segment the PCA time series into distinct behavioral states. 
Here, we compared ARHMMs and CP-\method with data from three mice. 
We used the first 35,949 timesteps of each recording, which were collected at 30Hz resolution. 
We used $H = 50$ discrete states and fitted ARHMMs and CP-\method models with varying lags and ranks.

The likelihood on a held-out validation set shows that the ARHMM overfitted quickly as the number of lags increased, while CP-\method was more robust to overfitting (Figure \ref{fig:moseq_result}B). We compared log-likelihoods of the best model (evaluated on the validation set) on a separate held-out test set and found that CP-\method consistently outperformed ARHMM across mice (Figure \ref{fig:moseq_result}C).

We also investigated the quality of \method segmentations of the behavioral data~(Appendix~\ref{app:sec:mouse:additional}).  We found that the PCA trajectories upon transition into a discrete \method state were highly stereotyped, suggesting that \method segments the data into consistent behavioral states.  Furthermore, CP-\method used fewer discrete states than the ARHMM, suggesting that the ARHMM may have oversegmented and that CP-\method offers a more parsimonious description of the data.

\subsection{Modeling \textit{C. elegans} Neural Data}
\label{sec:results:ce}

Finally, we analyzed neural recordings of an immobilized \textit{C. elegans} worm from~\citet{kato2015global}.  SLDS have previously been used to capture the time-varying low-dimensional dynamics of the neural activity~\citep{linderman2019hierarchical, glaser2020recurrent}.  We compared SLDS, ARHMM, and CP-\method with 18 minutes of neural traces (recorded at 3Hz; $\sim$3200 timesteps) from one worm, in which 48 neurons were confidently identified.  The dataset also contains 7 manually identified state labels based on the neural activity.

We used $H=7$ discrete states and fitted SLDSs, ARHMMs, and CP-\method with varying lags and ranks (or continuous latent dimensions for SLDSs).  Following \citet{linderman2019hierarchical}, we searched for sets of hyperparameters that achieve $\sim$90\% explained variance on a held-out test dataset (see Appendix \ref{sec:celegans_appendix} for more details).  For ARHMMs and CP-\method, we chose a larger lag ($L=9$, equivalent to 3 seconds) to examine the long-timescale correlations among the neurons.

We find that \method can perform as well as SLDSs and ARHMMs in terms of held-out explained variance ratio (a metric used by previous work~\citep{linderman2019hierarchical}).  As expected, we find that CP-\method can achieve these results with far fewer parameters than ARHMMs, and with a parameter count closer to SLDS than ARHMM (as more continuous latent states were required in an SLDS to achieve $\sim$90\% explained variance; see Appendix \ref{sec:celegans_appendix}).  Figure \ref{fig:celegans_analysis}A shows that \method, SLDS and ARHMM produce similar segmentations to the given labels, as evidenced by the confusion matrix having high entries on the leading diagonal (Figure \ref{fig:celegans_analysis}B and Appendix \ref{sec:celegans_appendix}).

Figure \ref{fig:celegans_analysis}C shows the one-dimensional autoregressive filters learned by CP-\method, defined as
%$\sum_{k=1}^{D} u_{ik} v_{jk} \mbw_{:k}$ 
$\sum_{i=1}^{D_1} \sum_{j=1}^{D_2} \sum_{k=1}^{D_3} g_{ijk}^{(h)} \, u_{pi}^{(h)} v_{qj}^{(h)} \mbw_{:k}^{(h)}$
for neurons $p$ and $q$.
We see that neurons believed to be involved in particular behavioral states have high weights in the filter (e.g., \texttt{SMDV} during the ``Ventral Turn'' state and \texttt{SMDD} during the ``Dorsal Turn'' state~\citep{linderman2019hierarchical, kato2015global, gray2005circuit, kaplan2020nested, yeon2018sensory}).  This highlights how switching autoregressive models can reveal state-dependent functional interactions between neurons (or observed states more generally).  
In Appendix \ref{sec:celegans_appendix}, we show the autoregressive filters learned by an ARHMM, an SLDS, and a generalized linear model (GLM), a method commonly used to model inter-neuronal interactions~\citep{pillow2008spatio}. 
Interestingly, the GLM does not find many strong functional interactions between neurons, likely because it is averaging over many unique discrete states.  In addition to its advantages in parameter efficiency and estimation, SALT thus provides a novel method for finding changing functional interactions across neurons at multiple timescales.

\begin{figure}[!t]
\begin{center}
	\includegraphics[width=\textwidth]{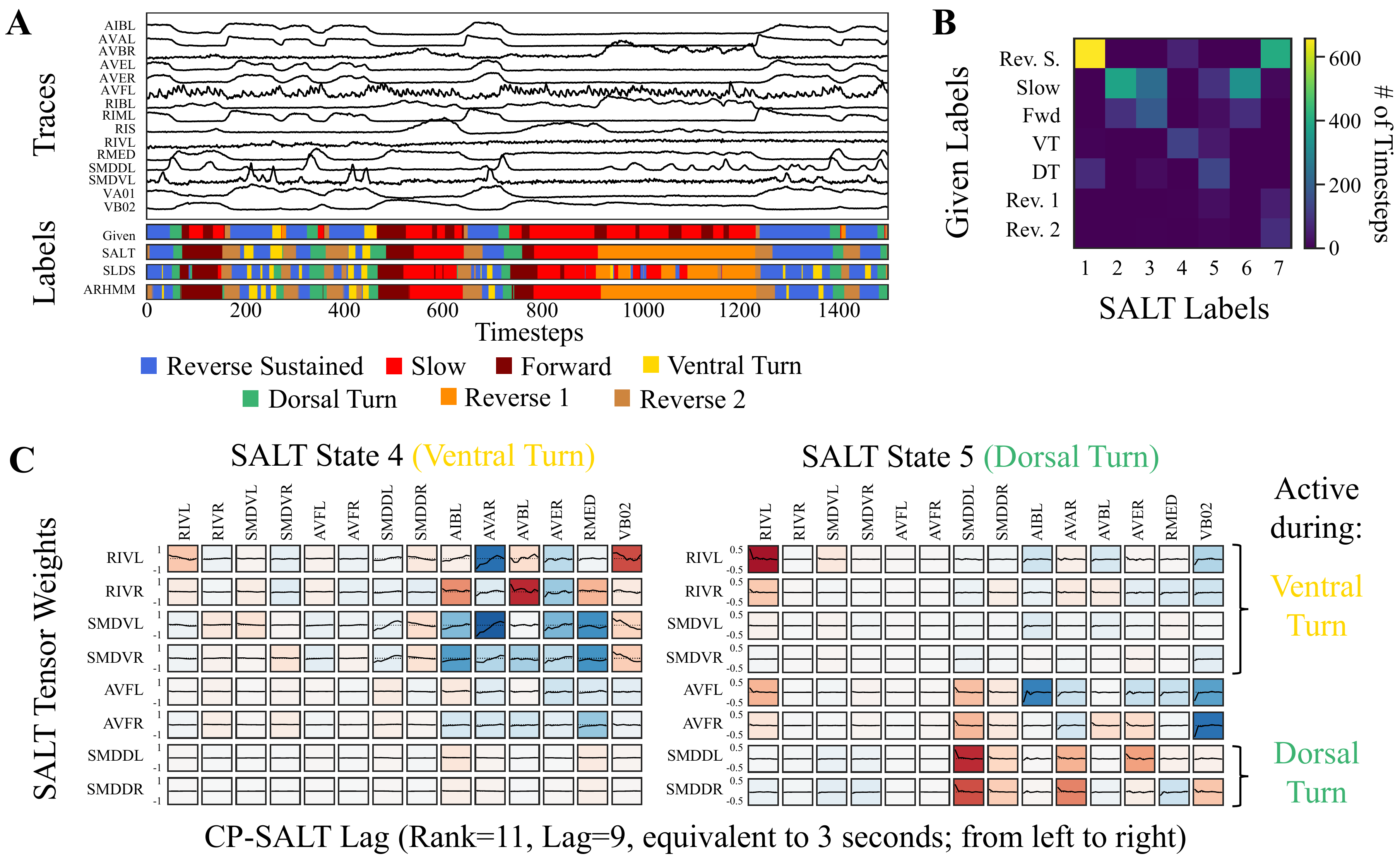}
    \vspace*{-0.2cm}
    \caption{\textbf{CP-\method provides good segmentations of \textit{C. elegans} neural data, and inferred low-rank tensors give insights into temporal dependencies among neurons in each discrete state}:  \textbf{(A)} Example data with manually generated labels (Given), as well as segmentations generated by \method, SLDS and ARHMM models. Learned states are colored based on the permutation of states that best matches given labels. All methods produce comparable segmentations, with high agreement with the given labels.  \textbf{(B)} Confusion matrix of \method-generated labels.  \textbf{(C)} One-dimensional autoregressive filters learned in two states by \method (identified as ventral and dorsal turns).  Colors indicate the area under curve (red is positive; blue is negative).  The first four rows are neurons known to mediate ventral turns, while the last two rows mediate dorsal turns~\citep{kato2015global, gray2005circuit, yeon2018sensory}.  These known behavior-tuned neurons generally have larger magnitude autoregressive filters.  Interestingly, \texttt{AVFL} and \texttt{AVFR} also have large filters for dorsal turns. These neurons do not have a well-known function.  However, they are associated with motor neurons, and so may simultaneously activate due to factors that co-occur with turning.  This highlights how \method may be used for proposing novel relationships in systems. }
    \label{fig:celegans_analysis}
\end{center}
\vspace*{-0.2cm}
\end{figure}

\section{Discussion}
We introduce switching autoregressive low-rank tensor (\method) models: a novel model class that parameterizes the autoregressive tensors of an ARHMM with a low-rank factorization.  
This constraint allows \method to model time-series data with fewer parameters than ARHMMs and with simpler estimation procedures than SLDSs.
We also make theoretical connections between low-rank tensor regressions and LDSs.
We then demonstrate, with both synthetic and real datasets, that SALT offers both efficiency and interpretability, striking an advantageous balance between the ARHMM and SLDS.
Moreover, \method offers an enhanced ability to investigate the interactions across observations, such as neurons, across different timescales in a data-efficient manner.

\method could be extended in many ways.
For example, neural spike trains are often modeled with Poisson likelihoods instead of \method's Gaussian noise model.  In this case, the E-step would still be exact, but the M-step would no longer have closed-form coordinate updates.
Likewise, the discrete state transitions could be allowed to depend on the current observations, as in recurrent state space models~\citep{linderman2017bayesian}.
Altogether, \method offers simple and effective means of modeling and inference for complex, time-varying dynamical systems.

\acksection
This work was supported by grants from the Simons Collaboration on the Global Brain (SCGB 697092), the NIH (U19NS113201, R01NS113119, R01NS130789, and K99NS119787), the Sloan Foundation, and the Stanford Center for Human and Artificial Intelligence.
We thank Liam Paninski for his constructive feedback on the paper.
We also thank the members of the Linderman Lab for their support and feedback throughout the project.

\clearpage
\bibliographystyle{unsrtnat}
\bibliography{refs}
\clearpage
\appendix

{\Large \textbf{Supplementary Materials for: Switching Autoregressive Low-rank Tensor Models}}

\

\section*{Table of Contents}
-- \textbf{Appendix \ref{sec:update}}: \method Optimization via Tensor Regression.

-- \textbf{Appendix \ref{sec:lds_tr}}: \method approximates a (Switching) Linear Dynamical System.

-- \textbf{Appendix \ref{app:sec:sss}}: Single-subspace \method.

-- \textbf{Appendix \ref{app:sec:syn}}: Synthetic Data Experiments.

-- \textbf{Appendix \ref{app:sec:mouse}}: Modeling Mouse Behavior.

-- \textbf{Appendix \ref{sec:celegans_appendix}}: Modeling \textit{C. elegans} neural data.

% -- \textbf{Appendix \ref{app:sec:code}}: Code Availability.

\newpage

\section{\method Optimization via Tensor Regression}
\label{sec:update}

Let $\mby_t \in \reals^{N_1}$ be the $t$-th outputs and $\mbX_t \in \reals^{N_2 \times N_3}$ be the $t$-th inputs.
The regression weights are a tensor~$\beA \in \reals^{N_1 \times N_2 \times N_3}$, which we model via a Tucker decomposition,
\begin{align}
    \beA &= \sum_{i=1}^{D_1} \sum_{j=1}^{D_2} \sum_{k=1}^{D_3} g_{ijk} \, \mbu_{:i} \circ \mbv_{:j} \circ \mbw_{:k},
\end{align}
where $\mbu_i$, $\mbv_j$, and $\mbw_k$ are columns of the factor matrices~$\mbU \in \reals^{N_1 \times D_1}$,~$\mbV \in \reals^{N_2 \times D_2}$, and~$\mbW \in \reals^{N_3 \times D_3}$, respectively, and $g_{ijk}$ are entries in the core tensor~$\beG \in \reals^{D_1 \times D_2 \times D_3}$.

Consider the linear model,~${\mby_t \sim \cN(\beA \times_{2,3} \mbX_t, \mbQ)}$
where ${\beA \times_{2,3} \mbX_t}$ is defined using the Tucker decomposition of $\beA$ as,
\begin{align}
 \beA \times_{2,3} \mbX_t
 &= \beA_{(1)} \mathrm{vec}(\mbX_t) \\
 &= \mbU \beG_{(1)} (\mbV^\top \otimes \mbW^\top) \mathrm{vec}(\mbX_t) \\
 &= \mbU \beG_{(1)} \mathrm{vec}( \mbV^\top \mbX_t \mbW) \label{eq:U_form}
\end{align}
where $\beA_{(1)} \in \reals^{N_1 \times N_2 N_3}$ and $\beG_{(1)} \in \reals^{D_1 \times D_2 D_3}$ are mode-1 matricizations of the corresponding tensors.
\textit{Note that these equations assume that matricization and vectorization are performed in row-major order, as in Python but opposite to what is typically used in Wikipedia articles.}

\Cref{eq:U_form} can be written in multiple ways, and these equivalent forms will be useful for deriving the updates below. We have,
\begin{align}
\beA \times_{2,3} \mbX_t
 &= \mbU \beG_{(1)} (\mbI_{D_2} \otimes \mbW^\top \mbX_t^\top) \mathrm{vec}(\mbV^\top) \label{eq:V_form}\\
 &= \mbU \beG_{(1)} (\mbV^\top \mbX_t \otimes \mbI_{D_3}) \mathrm{vec}(\mbW) \label{eq:W_form} \\
 &= \left[ \mbU \otimes \mathrm{vec}(\mbV^\top \mbX_t \mbW) \right]  \mathrm{vec}(\beG) \label{eq:G_form}.
\end{align}

We minimize the negative log likelihood by coordinate descent.

\paragraph{Optimizing the output factors}
Let
\begin{align}
    \widetilde{\mbx}_t &= \beG_{(1)} \mathrm{vec}( \mbV^\top \mbX_t \mbW)
\end{align}
for fixed $\mbV$, $\mbW$, and $\beG$. The NLL as a function of $\mbU$ is,
\begin{align}
    \cL(\mbU) &= \frac{1}{2} \sum_t (\mby_t - \mbU \widetilde{\mbx}_t)^\top \mbQ^{-1} (\mby_t - \mbU \widetilde{\mbx}_t).
\end{align}
This is a standard least squares problem with solution
\begin{align}
    \mbU^\star &= \left(\sum_t \mby_t \widetilde{\mbx}_t^\top \right) \left(\sum_t \widetilde{\mbx}_t \widetilde{\mbx}_t^\top \right)^{-1}.
\end{align}

\paragraph{Optimizing the core tensors}
Let $\widetilde{\mbX}_t = \mbU \otimes \mathrm{vec}(\mbV^\top \mbX_t \mbW) \in \reals^{N_1 \times D_1 D_2 D_3}$
denote the coefficient on $\mathrm{vec}(\beG)$ in~\cref{eq:G_form}. 
The NLL as a function of $\mbg = \mathrm{vec}(\beG)$ is,
\begin{align}
    \cL(\mbg) &= \frac{1}{2} \sum_t (\mby_t - \widetilde{\mbX}_t \mbg)^\top \mbQ^{-1} (\mby_t - \widetilde{\mbX}_t \mbg).
\end{align}
The minimizer of this quadratic form is,
\begin{align}
    \mbg^\star &= \left(\sum_t \widetilde{\mbX}_t^\top \mbQ^{-1} \widetilde{\mbX}_t \right)^{-1} \left(\sum_t \widetilde{\mbX}_t^\top \mbQ^{-1} \mby_t \right)
\end{align}

\paragraph{Optimizing the input factors}
Let
\begin{align}
 \widetilde{\mbX}_t
 &= \mbU \beG_{(1)} (\mbI_{D_2} \otimes \mbW^\top \mbX_t^\top)
\end{align}
for fixed $\mbU$, $\mbW$, and $\beG$. The NLL as a function of $\mbv = \mathrm{vec}(\mbV^\top)$ is,
\begin{align}
    \cL(\mbv) &= \frac{1}{2} \sum_t (\mby_t - \widetilde{\mbX}_t \mbv)^\top \mbQ^{-1} (\mby_t - \widetilde{\mbX}_t \mbv).
\end{align}
The minimizer of this quadratic form is,
\begin{align}
    \mbv^\star &= \left(\sum_t \widetilde{\mbX}_t^\top \mbQ^{-1} \widetilde{\mbX}_t \right)^{-1} \left(\sum_t \widetilde{\mbX}_t^\top \mbQ^{-1} \mby_t \right)
\end{align}

\paragraph{Optimizing the lag factors}
Let
\begin{align}
 \widetilde{\mbX}_t
 &= \mbU \beG_{(1)} (\mbV^\top \mbX_t \otimes \mbI_{D_3})
\end{align}
for fixed $\mbU$, $\mbV$, and $\beG$. The NLL as a function of $\mbw = \mathrm{vec}(\mbW)$ is,
\begin{align}
    \cL(\mbw) &= \frac{1}{2} \sum_t (\mby_t - \widetilde{\mbX}_t \mbw)^\top \mbQ^{-1} (\mby_t - \widetilde{\mbX}_t \mbw).
\end{align}
The minimizer of this quadratic form is,
\begin{align}
    \mbw^\star &= \left(\sum_t \widetilde{\mbX}_t^\top \mbQ^{-1} \widetilde{\mbX}_t \right)^{-1} \left(\sum_t \widetilde{\mbX}_t^\top \mbQ^{-1} \mby_t \right)
\end{align}

\paragraph{Multiple discrete states}
If we have discrete states $z_t \in \{1,\ldots,H\}$ and each state has its own parameters~$(\beG^{(h)}, \mbU^{(h)}, \mbV^{(h)}, \mbW^{(h)}, \mbQ^{(h)})$, then letting $\omega_{t}^{(h)} = \bbE[z_t = h]$ denote the weights from the E-step, the summations in coordinate updates are weighted by $\omega_{t}^{(h)}$.
For example, the coordinate update for the core tensors becomes,
\begin{align}
    \mbg^{(h) \star} &= \left(\sum_t \omega_{t}^{(h)} \widetilde{\mbX}_{t}^{(h)\top} \mbQ^{(h)-1} \widetilde{\mbX}_{t}^{(h)} \right)^{-1} \left(\sum_t \omega_{t}^{(h)} \widetilde{\mbX}_{t}^{(h) \top} \mbQ^{(h)-1} \mby_t \right)
\end{align}

\newpage

\section{\method approximates a (Switching) Linear Dynamical System}
\label{sec:lds_tr}
We now re-state and provide a full proof for Proposition \ref{prop:lds}.  

\setcounter{proposition}{0}
\begin{proposition}[Low-Rank Tensor Autoregressions Approximate Stable Linear Dynamical Systems]
    Consider a stable linear time-invariant Gaussian dynamical system.  We define the steady-state Kalman gain matrix as $\mbK = \lim_{t \rightarrow \infty} \mbK_t$, and $\mb\Gamma = \mbA(\mbI - \mbK\mbC)$.  The matrix $\mb\Gamma \in \reals^{D \times D}$ has eigenvalues $\lambda_1, \ldots, \lambda_D$. Let $\lambda_{\mathsf{max}} = \max_d |\lambda_d|$; for a stable LDS, $\lambda_{\mathsf{max}} < 1$~\emph{\citep{davis1985control}}. Let $n$ denote the number of real eigenvalues and $m$ the number of complex conjugate pairs. Let $\hat{\mby}_t^{(\mathsf{LDS})} = \mathbb{E}[\mby_t \mid \mby_{<t}]$ denote the predictive mean under a steady-state LDS, and $\hat{\mby}_t^{(\mathsf{SALT})}$ the predictive mean under a SALT model. An order-$L$ Tucker-SALT model with rank $n + 2m$, or a CP-SALT model with rank $n + 3m$, can approximate the predictive mean of the steady-state LDS with error $\|\hat{\mby}_t^{(\mathsf{LDS})} - \hat{\mby}_t^{(\mathsf{SALT})}\|_\infty = \cO(\lambda_{\mathsf{max}}^L)$.
\end{proposition}

\begin{proof}
A stationary linear dynamical system (LDS) is defined as follows:
\begin{align}
    \mbx_t &= \mbA \mbx_{t-1} + \mbb + \mb\epsilon_t \\
    \mby_t &= \mbC \mbx_{t} + \mbd + \mb\delta_t
\end{align}
where $\mby_t \in \reals^{N}$ is the $t$-th observation, $\mbx_t \in \reals^{D}$ is the $t$-th hidden state, $\mb\epsilon_t \overset{\text{i.i.d.}}{\sim} \cN(\mb0, \mbQ)$, $\mb\delta_t \overset{\text{i.i.d.}}{\sim} \cN(\mb0, \mbR)$, and $\mb\theta=(\mbA, \mbb, \mbQ, \mbC, \mbd, \mbR)$ are the parameters of the LDS.

Following the notation of \citet{murphy2012machine}, the one-step-ahead posterior predictive distribution for the observations of the LDS defined above can be expressed as:
\begin{align}
    p(\mby_t | \mby_{1:t-1}) 
	&= \cN(\mbC \mb\mu_{t|t-1} + \mbd, \mbC \mb\Sigma_{t|t-1} \mbC^T + \mbR)
\end{align}
where
\begin{align}
	\mb\mu_{t|t-1} &= \mbA \mb\mu_{t-1} + \mbb \label{equ:kalman_1}\\
	\mb\mu_{t} &= \mb\mu_{t|t-1} + \mbK_t \mbr_t \\
	\mb\Sigma_{t|t-1} &= \mbA \mb\Sigma_{t-1} \mbA^T + \mbQ \\
	\mb\Sigma_{t} &= (\mbI - \mbK_t \mbC) \mb\Sigma_{t|t-1} \\
	p(\mbx_1) &= \cN(\mbx_1 \mid \mb\mu_{1|0}, \mb\Sigma_{1|0}) \\
	\mbK_t &= (\mb\Sigma_{t|t-1}^{-1} + \mbC^T \mbR \mbC)^{-1} \mbC^T \mbR^{-1} \\
	\mbr_t &= \mby_t - \mbC \mb\mu_{t|t-1} - \mbd.  \label{equ:kalman_f}
\end{align}

We can then expand the mean $\mbC \mb\mu_{t|t-1} + \mbd$ as follows:
\begin{align}
    \mbC \mb\mu_{t|t-1} + \mbd &= \mbC\sum_{l=1}^{t-1} \mb\Gamma_l \mbA \mbK_{t-l} \mby_{t-l} 
    +\mbC \sum_{l=1}^{t-1} \mb\Gamma_l (\mbb - \mbA \mbK_{t-l} \mbd) + \mbd
\end{align}
where
\begin{align}
	\mb\Gamma_{l} &= {\prod_{i=1}^{l-1}} \mbA (\mbI-\mbK_{t-i}\mbC) \quad \mathrm{for}\quad  l \in \left\lbrace 2, 3, \ldots \right\rbrace, \\
    \mb\Gamma_{1} &= \mbI.
\end{align}

Theorem 3.3.3 of \citet{davis1985control} (reproduced with our notation below) states that for a stabilizable and detectable system, the $\text{lim}_{t \rightarrow \infty} \mb\Sigma_{t|t-1} = \mb\Sigma$, where $\mb\Sigma$ is the unique solution of the discrete algebraic Riccati equation 
\begin{equation}
    \mb\Sigma = \mbA \mb\Sigma \mbA^T - \mbA \mb\Sigma \mbC^T (\mbC \mb\Sigma \mbC^T + \mbR)^{-1} \mbC \mb\Sigma \mbA^T + \mbQ. \label{equ:dari}
\end{equation}  As we are considering stable autonomous LDSs here, the system is stabilizable and detectable, as all unobservable states are themselves stable~\citep{davis1985control, katayama2005subspace}

\renewcommand\thetheorem{3.3.3}
\begin{theorem}[Reproduced from~\citet{davis1985control}, updated to our notation and context] 
The theorem has two parts.
\begin{enumerate}[label=(\alph*)]
\item If the pair $(\mbA, \mbC)$ is detectable then there exists at least one non-negative solution, $\mbSigma$, to the discrete algebraic Riccati equation \eqref{equ:dari}.\\
\item If the pair $(\mbA, \mbC)$ is stabilizable then this solution $\mb\Sigma$ is unique, and $\mb\Sigma_{t|t-1} \rightarrow \mb\Sigma$ as $t \rightarrow \infty$, where $\mb\Sigma_{t|t-1}$ is the sequence generated by \eqref{equ:kalman_1}-\eqref{equ:kalman_f} with arbitrary initial covariance $\mb\Sigma_0$.  
Then, the matrix $\mb\Gamma = \mbA (\mbI - \mbK \mbC)$ is stable, where $\mbK$ is the Kalman gain corresponding to~$\mb\Sigma$; i.e.,
\begin{equation}
    \mbK = (\mb\Sigma^{-1} + \mbC^T \mbR \mbC)^{-1} \mbC^T \mbR^{-1}
\end{equation}
\end{enumerate}
\end{theorem}

\begin{proof}
See \citet{davis1985control}.  Note that \citet{davis1985control} define the Kalman gain as~$\mbA\mbK$.
\end{proof}

The convergence of the Kalman gain also implies that each term in the sequence $\mb\Gamma_l$ converges to
\begin{align}
	\mb\Gamma_{l} &= {\prod_{i=1}^{l-1}} \mbA (\mbI-\mbK \mbC) = (\mbA (\mbI - \mbK \mbC))^{l-1} = \mb\Gamma^{l-1},
\end{align}
where, concretely, we define $\mb\Gamma = \mbA (\mbI - \mbK \mbC)$.  We can therefore make the following substitution and approximation
\begin{align}
    \mbC \mb\mu_{t|t-1} + \mbd & \stackrel{\lim t \rightarrow \infty}{=} \mbC\sum_{l=1}^{t-1} \mb\Gamma^l \mbA \mbK \mby_{t-l} +\mbC \sum_{l=1}^{t-1} \mb\Gamma^l (\mbb - \mbA \mbK \mbd) + \mbd  \label{app:equ:seq_full} \\
    &= \mbC\sum_{l=1}^{L} \mb\Gamma^l \mbA \mbK \mby_{t-l} + \mbC \sum_{l=1}^{L} \mb\Gamma^l (\mbb - \mbA \mbK \mbd) + \mbd + \textcolor{blue}{\sum_{l=L+1}^{\infty}\mathcal{F} \left( \mb\Gamma^{l} \right)} \label{app:equ:seq} \\ 
    &\approx \mbC\sum_{l=1}^{L} \mb\Gamma^l \mbA \mbK \mby_{t-l} + \mbC \sum_{l=1}^{L} \mb\Gamma^l (\mbb - \mbA \mbK \mbd) + \mbd \label{app:equ:truncated_kalman_form}
\end{align}
The approximation is introduced as a result of truncating the sequence to consider just the ``first'' $L$ terms, and discarding the higher-order terms (indicated in \textcolor{blue}{blue}).  It is important to note that each term in \eqref{app:equ:seq_full} is the sum of a geometric sequence multiplied elementwise with $\mby_t$.  

There are two components we prove from here.  First, we derive an element-wise bound on the error introduced by the truncation, and verify that under the conditions outlined that the bound decays monotonically in $L$.  We then show that Tucker and CP decompositions can represent the truncated summations in \eqref{app:equ:truncated_kalman_form}, and derive the minimum rank required for this representation to be exact.

\paragraph{Bounding The Error Term}
We first rearrange the truncated terms in \eqref{app:equ:seq_full}, where we define $\mbx_l \triangleq \mbA \mbK \mby_{t-l} + \mbb - \mbA \mbK \mbd$
\begin{align}
    \textcolor{blue}{\sum_{l=L+1}^{\infty}\mathcal{F} \left( \mb\Gamma^{l} \right)} &= \mbC\sum_{l=L+1}^{\infty} \mb\Gamma^l \mbA \mbK \mby_{t-l} + \mbC \sum_{l=L+1}^{\infty} \mb\Gamma^l (\mbb - \mbA \mbK \mbd) + \mbd , \\
    &= \sum_{l=L+1}^{\infty} \mbC \mb\Gamma^{l}\mbx_l, \\
    &= \sum_{l=L+1}^{\infty} \mbC \mbE \mb\Lambda^{l-1} \mbE^{-1} \mbx_l, \\
    &= \sum_{l=L+1}^{\infty} \mbP \mb\Lambda^{l-1} \mbq_{l},
\end{align}
where $\mbE \mb\Lambda \mbE^{-1}$ is the eigendecomposition of $\mb\Gamma$, $\mbP \triangleq \mbC \mbE$, and $\mbq_{l} \triangleq \mbE^{-1} \mbx_l$.  We now consider the infinity-norm of the error, and apply the triangle and Cauchy-Schwartz inequalities.
We can write the bound on the  as
\begin{align}
    \epsilon &= \left| \left( \textcolor{blue}{\sum_{l=L+1}^{\infty}\mathcal{F} \left( \mb\Gamma^{l} \right)} \right)_n \right|, \quad \mathrm{where} \quad n = \argmax_k \left| \left( \textcolor{blue}{\sum_{l=L+1}^{\infty}\mathcal{F} \left( \mb\Gamma^{l} \right)} \right)_k \right| \\ 
    &= \left| \sum_{l=L+1}^{\infty} \sum_{d=1}^D p_{nd} \lambda_d^{l-1} q_{l, d} \right|, \\
    &\leq  \sum_{l=L+1}^{\infty} \sum_{d=1}^D \left| p_{nd} \right| \left| \lambda_d^{l-1} \right| \left| q_{l, d} \right| .
\end{align}
Upper bounding the absolute magnitude of $q_{l, d}$ by $W$ provides a further upper bound, which we can then rearrange
\begin{align}
    \epsilon &\leq W \sum_{l=L+1}^{\infty} \sum_{d=1}^D \left| p_{nd} \right| \left| \lambda_d^{l-1} \right|, \\
    &= W \sum_{d=1}^D \left| p_{nd} \right| \sum_{l=L+1}^{\infty} \left| \lambda_d^{l-1} \right|.
\end{align}
The first two terms are constant, and hence the upper bound is determined by the sum of the of the $l^{\mathrm{th}}$ power of the eigenvalues.  We can again bound this sum by setting all eigenvalues equal to the magnitude of the eigenvalue with the maximum magnitude (spectral norm), denoted $\lambda_{max}$:
\begin{align}
    \epsilon &\leq W \sum_{d=1}^D \left| p_{nd} \right| \sum_{l=L+1}^{\infty}\lambda_{\mathrm{max}}^{l-1}, 
\end{align}
where these second summation is not a function of $d$, and $W \sum_{d=1}^D \left| p_{nd} \right|$ is constant.  This summation is a truncated geometric sequence.  Invoking Theorem 3.3.3 of \citet{davis1985control} again, the matrix $\mb\Gamma$ has only stable eigenvalues, and hence $\lambda_{\mathrm{max}} < 1$.  Therefore the sequence sum will converge to
\begin{align}
    \sum_{l=L+1}^{\infty}  \lambda_{\mathrm{max}}^{l-1} = \frac{\lambda_{\mathrm{max}}^L}{1 - \lambda_{\mathrm{max}}}.
\end{align}
Rearranging again, we see that the absolute error on the $n^{\mathrm{th}}$ element of $\mby_t$ is therefore bounded according to a power of the spectral norm
\begin{align}
    \epsilon &\leq W \sum_{d=1}^D \left| p_{nd} \right| \frac{\lambda_{\mathrm{max}}^L}{1 - \lambda_{\mathrm{max}}}, \\
    &= \mathcal{O}\left( \lambda_{\mathrm{max}}^L \right).
\end{align}
More specifically, for a stable linear time-invariant dynamical system, and where $\mbq$ --- and hence $\mby$ --- is bounded, then the bound on the error incurred reduces exponentially in the length of the window $L$.  Furthermore, this error bound will reduce faster for systems with a lower spectral norm.

\paragraph{Diagonalizing the System}
We first transform $\mb\Gamma$ into real modal form, defined as $\mbE \mb\Lambda \mbE^{-1}$, where $\mbE$ and $\mb\Lambda$ are the eigenvectors and diagonal matrix of eigenvalues of $\mb\Gamma$.  Letting $\mb\Gamma$ have $n$ real eigenvalues and $m$ pairs of complex eigenvalues (i.e., $n + 2m=D$), we can express $\mbE$, $\mb\Lambda$, and $\mbE^{-1}$ as:
\begin{align}  
\mbE = 
\setlength\arraycolsep{2pt}
\left[
  \begin{array}{*8{c}}
    \mba_{1}     & \ldots & \mba_{n}    & \mbb_1      & \mbc_1      & \ldots & \mbb_m      & \mbc_m      \\
  \end{array}
\right]
\end{align}  

\begin{align}  
\mb\Lambda = 
\setlength\arraycolsep{2pt}
\left[
  \begin{array}{*8{c}}
    \lambda_1 &        &           &          &        &         &        &\\
              & \ddots &           &          &        &         &        &\\
              &        & \lambda_n &          &        &         &        &\\
              &        &           & \sigma_1 &\omega_1&         &        &\\
              &        &           &-\omega_1 &\sigma_1&         &        &\\
              &        &           &          &        & \ddots  &        &\\
              &        &           &          &        &         & \sigma_m &\omega_m\\
              &        &           &          &        &         & -\omega_m &\sigma_m\\
  \end{array}
\right]
\end{align}
\begin{align} 
\mbE^{-1} = 
\setlength\arraycolsep{2pt}
\left[
  \begin{array}{*1{c}}
    \mbd^{T}_{1}  \\
     \vdots  \\
     \mbd^{T}_{n}  \\
     \mbe^{T}_{1}  \\
     \mb f^{T}_{1}  \\
     \vdots      \\
     \mbe^{T}_{m}  \\
     \mb f^{T}_{m} 
  \end{array}
\right]
\end{align}
where $\mba_1 \dots \mba_n$ are the right eigenvectors corresponding to $n$ real eigenvalues $\lambda_1 \dots \lambda_n$, and $\mbb_i$ and $\mbc_i$ are the real and imaginary parts of the eigenvector corresponding to the complex eigenvalue $\sigma_i+j\omega_i$. Note that
\begin{align}
	\mb\Gamma^{l} = (\mbA (\mbI-\mbK \mbC))^{l-1} = \mbE \mb\Lambda^{l-1} \mbE^{-1}
\end{align}
The $l^{th}$ power of $\mb\Lambda$, $\mb\Lambda^{l}$, where $l \geq 0$, can be expressed as:

\begin{align}  
\mb\Lambda^{l} = 
\setlength\arraycolsep{2pt}
\left[
  \begin{array}{*8{c}}
    \lambda_1^l &        &           &          &        &         &        &\\
              & \ddots &           &          &        &         &        &\\
              &        & \lambda_n^l &          &        &         &        &\\
              &        &           & \sigma_{1,l} &\omega_{1,l}&         &        &\\
              &        &           &-\omega_{1,l} &\sigma_{1,l}&         &        &\\
              &        &           &          &        & \ddots  &        &\\
              &        &           &          &        &         & \sigma_{m,l} &\omega_{m,l}\\
              &        &           &          &        &         & -\omega_{m,l} &\sigma_{m,l}\\
  \end{array}
\right]
\end{align}
where $\sigma_{i,l}=\sigma_{i,l-1}^2-\omega_{i,l-1}^2$, $\omega_{i,l}=2\sigma_{i,l-1}\omega_{i,l-1}$ for $l \geq 2$, $\sigma_{i,1}=\sigma_i$, $\omega_{i,1}=\omega_i$, $\sigma_{i,0}=1$, and $\omega_{i,0}=0$.

\paragraph{Tucker Tensor Regression}
Let $\beH \in \reals^{D \times D \times L}$ be a three-way tensor, whose $l^{th}$ frontal slice $\mbH_{::l} = \mbLambda^{l-1}$.
Let $\beG \in \reals^{D \times D \times D}$ be a three-way tensor, whose entry $g_{ijk} = \mathbbm{1}_{i=j=k}$ for $1 \leq k \leq n$, and $g_{ijk} = (-1)^{\mathbbm{1}_{i+1=j=k+1}} \mathbbm{1}_{(i=j=k) \lor (i-1=j-1=k) \lor (i=j+1=k+1) \lor (i+1=j=k+1)}$ for $k \in \{n+1, n+3, \dots, n+2m-1\}$.
Let $\mbW \in \reals^{L \times D}$ be a matrix, whose entry $w_{lk} = \lambda_k^{l-1}$ for $1 \leq k \leq n$, $w_{lk} = \sigma_{k,l-1}$ for $k \in \{n+1, n+3, \dots, n+2m-1\}$, and $w_{lk} = -\omega_{k,l-1}$ for $k \in \{n+2, n+4, \dots, n+2m\}$.
We can then decompose $\beH$ into $\beG \in \reals^{D \times D \times D}$ and $\mbW \in \reals^{L \times D}$ such that $\beH=\beG \times_{3} \mbW$ (Figure \ref{figsupp:Tucker-SALT}).

With $\mbV = (\mbE^{-1} \mbA \mbK)^T$, $\mbU = \mbC\mbE$, $\mbm=\mbC \sum_{l=1}^{L} \mb\Gamma^l (\mbb - \mbA \mbK \mbd) + \mbd$, and $\mbX_t = \mby_{t-1:t-L}$, we can rearrange the mean to:
\begin{align}
    \mbC \mb\mu_{t|t-1} + \mbd &\approx \mbC\sum_{l=1}^{L} \mbE \mb\Lambda^{l-1} \mbE^{-1} \mbA \mbK \mby_{t-l} 
    +\mbC \sum_{l=1}^{L} \mb\Gamma^l (\mbb - \mbA \mbK \mbd) + \mbd \label{eq:tucker_form} \\
    &= \mbU\sum_{l=1}^{L} \mbH_{::l} \mbV^T \mby_{t-l} + \mbm \\
    &= \mbU\sum_{l=1}^{L} (\beG \bar{\times}_3 \mbw_l) \mbV^T \mby_{t-l} + \mbm \\
    &= \mbU\sum_{l=1}^{L} ((\beG \times_{2} \mbV) \bar{\times}_3 \mbw_l) \mby_{t-l} + \mbm \\
    &= \mbU\sum_{l=1}^{L} \sum_{j=1}^{D} \sum_{k=1}^{D} \mbg_{:jk} \circ \mbv_{:j} (w_{lk} \mby_{t-l}) + \mbm \\
    &= \mbU \sum_{j=1}^{D} \sum_{k=1}^{D} \mbg_{:jk} (\mbv_{:j}^\top \mbX_t \mbw_{:k}) + \mbm \\
    &= \sum_{i=1}^{D} \sum_{j=1}^{D} \sum_{k=1}^{D} \mbu_{:i} g_{ijk} (\mbv_{:j}^\top \mbX_t \mbw_{:k}) + \mbm \\
    &= \left[\sum_{i=1}^{n+2m} \sum_{j=1}^{n+2m} \sum_{k=1}^{n+2m} g_{ijk} \mbu_{:i} \circ \mbv_{:j} \circ \mbw_{:k} \right] \times_{2,3} \mbX_t  + \mbm
\end{align}

\begin{figure*}[!t]
\begin{center}
	\includegraphics[width=\textwidth]{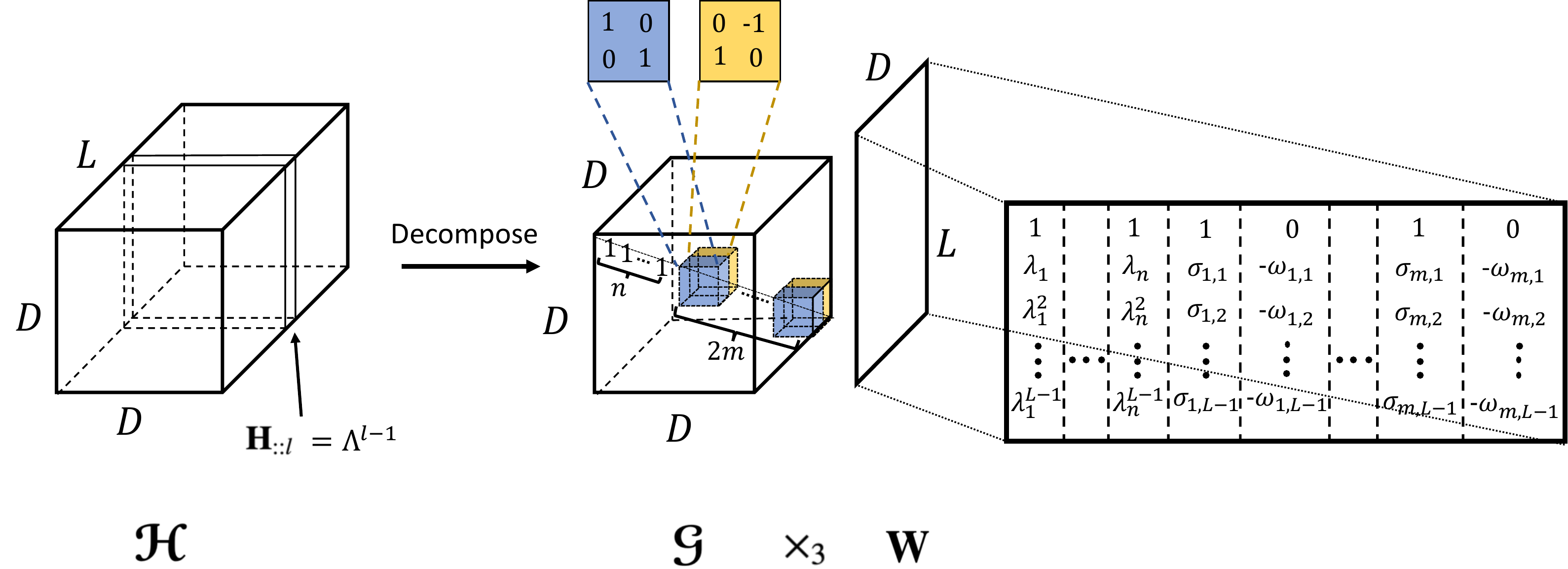}
    \caption{\textbf{Decomposition of $\beH$ into $\beG$ and $\mbW$ such that $\beH = \beG \times_3 \mbW$:} Given an LDS whose $\mbA(\mbI - \mbK \mbC)$ has $n$ real eigenvalues and $m$ pairs of complex eigenvalues, this decomposition illustrates how Tucker-\method can approximate the LDS well with rank $n+2m$.}
    \label{figsupp:Tucker-SALT}
\end{center}
\end{figure*}

\paragraph{CP Tensor Regression}

By rearranging $\mbE, \mb\Lambda^l,$ and $\mbE^{-1}$ into $\mbJ$, $\mbP_l$, and $\mbS$ respectively as follows:
\begin{align}  
\mbJ = 
\setlength\arraycolsep{2pt}
\left[
  \begin{array}{cccccccccc}
    \mba_{1}  & \ldots & \mba_{n} & \mbb_1+\mbc_1  & \mbb_1  & \mbc_1   & \ldots   & \mbb_m+\mbc_m & \mbb_m   & \mbc_m \\
  \end{array}
\right]
\end{align}  
\begin{align}  
\mbP_l = 
\setlength\arraycolsep{2pt}
\left[
  \begin{array}{cccccccccc}
    \lambda_1^l &        &           &          &         &        &        &        &         &\\
              & \ddots &           &          &         &        &        &        &         &\\
              &        & \lambda_n^l &          &         &        &        &        &         &\\
              &        &           & \sigma_{1,l} &         &        &        &        &         &\\
              &        &           &          &\alpha_{1,l} &        &        &        &         &\\
              &        &           &          &         &\beta_{1,l} &        &        &         &\\
              &        &           &          &         &        & \ddots &        &         &\\
              &        &           &          &         &        &        &\sigma_{m,l}&         &\\
              &        &           &          &         &        &        &        &\alpha_{m,l} &\\
              &        &           &          &         &        &        &        &         &\beta_{m,l}\\
  \end{array}
\right]
\end{align}
\begin{align} 
\mbS = 
\setlength\arraycolsep{2pt}
\left[
  \begin{array}{*1{c}}
    \mbd^{T}_{1}  \\
     \vdots          \\
     \mbd^{T}_{n}  \\
     \mbe^{T}_{1}+\mb f^{T}_{1}  \\
     \mb f^{T}_{1}  \\
     \mb e^{T}_{1}  \\
     \vdots     \\
     \mbe^{T}_{m}+\mb f^{T}_{m}  \\
     \mb f^{T}_{m}  \\
     \mb e^{T}_{m} 
  \end{array}
\right]
\end{align}

where $\mbJ \in \reals^{D \times (n+3m)}$, $\mbP_l \in \reals^{(n+3m) \times (n+3m)}$, $\mbS \in \reals^{(n+3m) \times D}$, $\alpha_{i,l}=\omega_{i,l}-\sigma_{i,l}$, and $\beta_{i,l}=-\omega_{i,l}-\sigma_{i,l}$, we can diagonalize $(\mbA(\mbI - \mbK \mbC))^l$ as $\mbJ \mbP_l \mbS$.

Let $\mbV = (\mbS \mbA \mbK)^T$, $\mbU = \mbC\mbJ$, $\mbm=\mbC \sum_{l=1}^{L} \mb\Gamma^l (\mbb - \mbA \mbK \mbd) + \mbd$, and $\mbX_t = \mby_{t-1:t-L}$.
Let $\mbW \in \mathbb{R}^{L \times (n+3m)}$ be a matrix, whose element in the $l^{th}$ row and $k^{th}$ column is $p_{l-1,kk}$ (i.e., the element in the $k^{th}$ row and $k^{th}$ column of $\mbP_{l-1}$), and $\beG \in \reals^{(n+3m) \times (n+3m) \times (n+3m)}$ be a superdiagonal 3-way tensor, where $g_{ijk}=\mathbbm{1}_{i=j=k}$.
We can then rearrange the mean to:
\begin{align}
    \mbC \mb\mu_{t|t-1} + \mbd & \approx \mbC\sum_{l=1}^{L} \mbE \mb\Lambda^{l-1} \mbE^{-1} \mbA \mbK \mby_{t-l} 
    + \mbC \sum_{l=1}^{L} \mb\Gamma^l (\mbb - \mbA \mbK \mbd) + \mbd \\
    & = \mbC\sum_{l=1}^{L}\mbJ \mbP_{l-1} \mbS \mbA \mbK \mby_{t-l} + \mbm \\
    & = \mbU\sum_{l=1}^{L} \mbP_{l-1} \mbV^\top \mby_{t-l} + \mbm \\
    & = \sum_{l=1}^{L} \sum_{i}^{n+3m} \sum_{j}^{n+3m} \sum_{k}^{n+3m} g_{ijk} \, \mbu_{:i} \circ \mbv_{:j} (p_{l-1,kk} \mby_{t-l}) + \mbm \\
    & = \sum_{i}^{n+3m} \sum_{j}^{n+3m} \sum_{k}^{n+3m} g_{ijk} \, \mbu_{:i} \circ \mbv_{:j} (\mbX_t \mbw_{:k}) + \mbm \\
    & = \left[\sum_{i=1}^{n+3m} \sum_{j=1}^{n+3m} \sum_{k=1}^{n+3m} g_{ijk} \, \mbu_{:i} \circ \mbv_{:j} \circ \mbw_{:k} \right] \times_{2,3} \mbX_t + \mbm
\end{align}
And so concludes the proof. 
\end{proof}

\newpage

\section{Single-subspace \method} 
\label{app:sec:sss}

Here we explicitly define the generative model of multi-subspace and single-subspace Tucker-\method and CP-\method.  Single-subspace \method is analogous to single-subspace SLDSs (also defined below), where certain emission parameters (e.g., $\mbC$, $\mbd$, and $\mbR$) are shared across discrete states.  This reduces the expressivity of the model, but also reduces the number of parameters in the model.  Note that both variants of all models have the same structure on the transition dynamics of $\mbz_t$.

\paragraph{Multi-subspace \method}
Note that the \method model defined in \eqref{equ:salt:likelihood} and \eqref{equ:salt:A_def} in the main text is a multi-subspace \method.  We repeat the definition here for ease of comparison.
\begin{align}
    \mby_t &\overset{\text{i.i.d.}}{\sim} \mathcal{N}\left( \left(  \sum_{i=1}^{D_1} \sum_{j=1}^{D_2} \sum_{k=1}^{D_3} g_{ijk}^{(z_t)} \, \mbu_{:i}^{(z_t)} \circ \mbv_{:j}^{(z_t)} \circ \mbw_{:k}^{(z_t)} \right) \times_{2,3} \mby_{t-1:t-L} + \mbb^{(z_t)} ,\pmb{\Sigma}^{(z_t)} \right),
\end{align}
$D_1 = D_2 = D_3 = D$ and $\beG$ is diagonal for CP-\method.

\paragraph{Single-subspace Tucker-\method}
In single-subspace methods, the output factors are shared across discrete states
\begin{align}
    \mby_t &\overset{\text{i.i.d.}}{\sim} \mathcal{N}\left(  \mbU \left( \mbm^{(z_t)} +  \left( \sum_{j=1}^{D_2} \sum_{k=1}^{D_3} \mbg_{:jk}^{(z_t)} \circ \mbv_{:j}^{(z_t)} \circ \mbw_{:k}^{(z_t)} \right) \times_{2,3} \mby_{t-1:t-L} \right) + \mbb ,\pmb{\Sigma}^{(z_t)} \right),
\end{align}
where $\mbm^{(z_t)} \in \reals^{D_1}$.

\paragraph{Single-subspace CP-\method}
Single-subspace CP-\method requires an extra tensor compared to Tucker-\method, as this tensor can no longer be absorbed in to the core tensor.
\begin{align}
    \mby_t &\overset{\text{i.i.d.}}{\sim} \mathcal{N}\left(  \mbU' \left( \mbm^{(z_t)} +  \mbP^{(z_t)} \left(\left( \sum_{j=1}^{D_2} \sum_{k=1}^{D_3} \mbg_{:jk}^{(z_t)} \circ \mbv_{:j}^{(z_t)} \circ \mbw_{:k}^{(z_t)} \right) \times_{2,3} \mby_{t-1:t-L} \right) \right) + \mbb ,\pmb{\Sigma}^{(z_t)} \right),
\end{align}
where $\mbU' \in \reals^{N \times D_1^{\prime}}$, $\mbP^{(z_t)} \in \reals^{D_1^{\prime} \times D_1}$, $\mbm^{(z_t)} \in \reals^{D_1^{\prime}}$, $D_1 = D_2 = D_3 = D$, and $\beG$ is diagonal.

\paragraph{Multi-subspace SLDS}
Multi-subspace SLDS is a much harder optimization problem, which we found was often numerically unstable.
We therefore do not consider multi-subspace SLDS in these experiments, but include its definition here for completeness
\begin{align}
    \mbx_t &\sim \cN\left(\mbA^{(z_t)} \mbx_{t-1} + \mbb^{(z_t)}, \, \mbQ^{(z_t)}\right),  \\
    \mby_t &\sim \cN\left(\mbC^{(z_t)} \mbx_t + \mbd^{(z_t)}, \, \mbR^{(z_t)}\right).
\end{align}

\paragraph{Single-subspace SLDS}
Single-subspace SLDS was used in all of our experiments, and is typically used in practice~\citep{petreska2011dynamical, linderman2017bayesian}
\begin{align}
    \mbx_t &\sim \cN\left(\mbA^{(z_t)} \mbx_{t-1} + \mbb^{(z_t)}, \, \mbQ^{(z_t)}\right),  \\
    \mby_t &\sim \cN\left(\mbC \mbx_t + \mbd, \, \mbR\right).
\end{align}

\newpage

\section{Synthetic Data Experiments}
\label{app:sec:syn}

\subsection{Extended Experiments for \Cref{prop:lds}}
\label{subsec:extended_theory_experiments}

In Section \ref{sec:results:lds} we showed that Proposition \ref{prop:lds} can accurately predict the required rank for CP- and Tucker-\method models.  We showed results for a single LDS for clarity.  We now extend this analysis across multiple random LDS and \method models.  We randomly sampled LDSs with latent dimensions ranging from 4 to 10, and observation dimensions ranging from 9 to 20. For each LDS, we fit 5 randomly initialized CP-\method and Tucker-\method models with $L=50$ lags.  We varied the rank of our fit \method models according to the rank predicted by \Cref{prop:lds}.  Specifically, we computed the estimated number of ranks for a given LDS, denoted $D^*$, and then fit \method models with $\left\lbrace D^*-2, D^*-1, D^*, D^*+1, D^*+2 \right\rbrace$ ranks.  According to Proposition \ref{prop:lds}, we would expect to see the reconstruction error of the autoregressive tensor be minimized, and for prediction accuracy to saturate, at $D=D^*$.  

\begin{figure}[!h]
    \centering
    \begin{subfigure}[b]{0.475\textwidth}
    	\includegraphics[width=\textwidth]{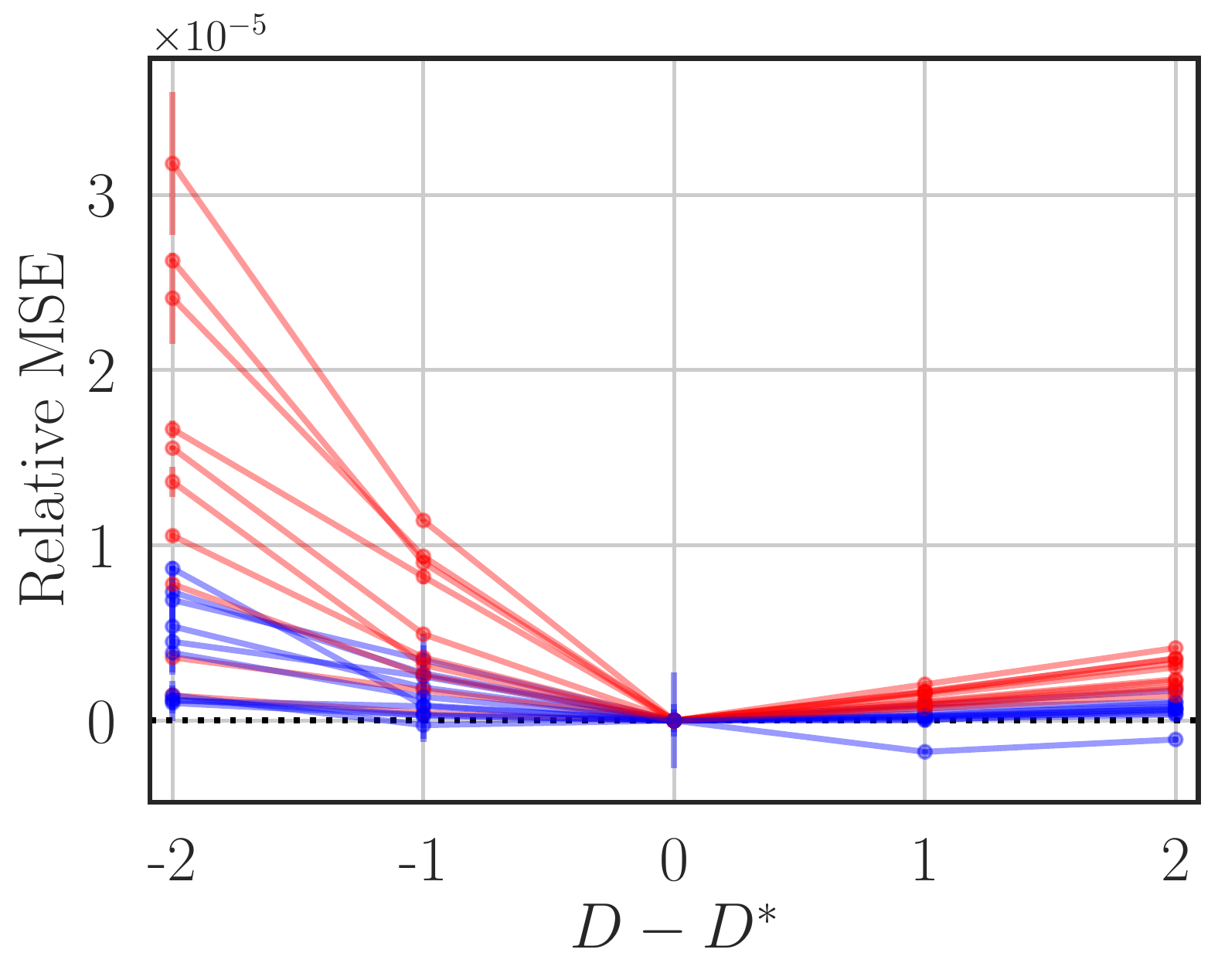}
        \caption{Normalized MSE of autoregressive tensor.}
        \label{figsupp:extended_theory_mse}
    \end{subfigure}%
    \hfill%
    \begin{subfigure}[b]{0.52\textwidth}
    	\includegraphics[width=\textwidth]{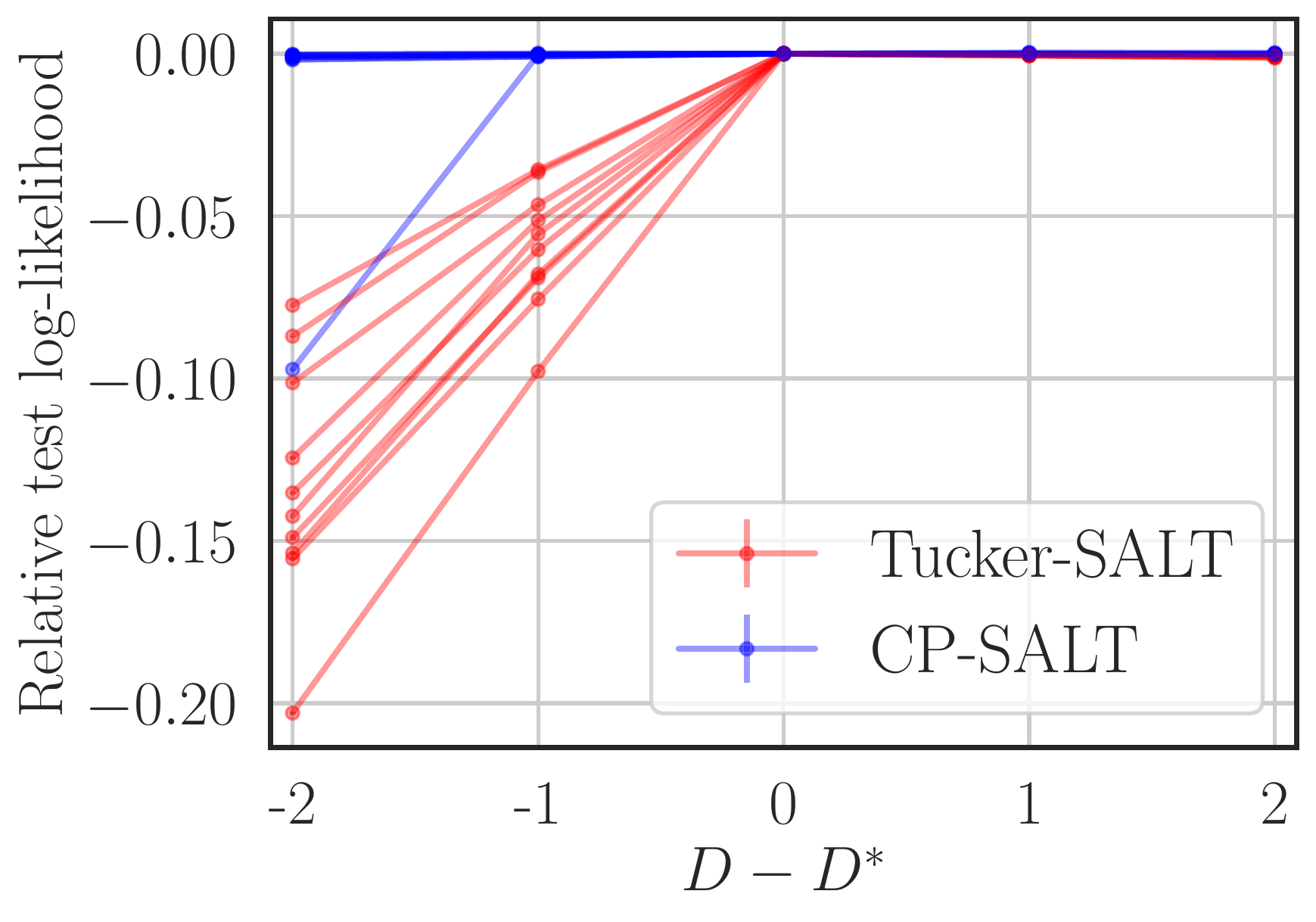}
        \caption{Normalized log-likelihood on held-out test set.}
        \label{figsupp:extended_theory_ll}
    \end{subfigure}
    \caption{Extended results examining Proposition \ref{prop:lds}.  Results are shown for the ability of \method to estimate ten randomly generated LDSs, using five \method repeats for each LDS.  MSEs (in panel A) and log-likelihoods (in panel B) are normalized by the mean MSE and mean log-likelihood of \method models trained with $D=D^*$. $D$ is the rank of the fit \method model, and $D^*$ is the necessary rank predicted by \Cref{prop:lds}. }
    \label{app:fig:lds}
\end{figure}

To analyze these model fits, we first computed the average mean squared error of the autoregressive tensor corresponding to the LDS simulation,  as a function of \method rank relative to the rank required by \Cref{prop:lds}.  We see, as predicted by Proposition \ref{prop:lds}, that error in the autoregressive tensor is nearly always minimized at $D^*$ (Figure \ref{app:fig:lds}A).  Tucker-\method was always minimized at $D^*$.  Some CP-\method fits have lower MSE at ranks other than predicted by \Cref{prop:lds}.  We believe this is due to local minima in the optimization. We next investigated the test log-likelihood as a function of the relative rank (Figure \ref{app:fig:lds}B). Interestingly, the test log-likelihood shows that Tucker-\method strongly requires the correct number of ranks for accurate prediction, but CP-\method can often perform well with fewer ranks than predicted (although still a comparable number of ranks to Tucker-\method). As in Figure~\ref{fig:lds_simulated}, these analyses empirically confirm \Cref{prop:lds}.

\subsection{Quantitative Performance: Synthetic Switching LDS Experiments}
\begin{figure}[!h]
    \centering
    \begin{subfigure}[b]{0.47\textwidth}
    	\includegraphics[width=\textwidth]{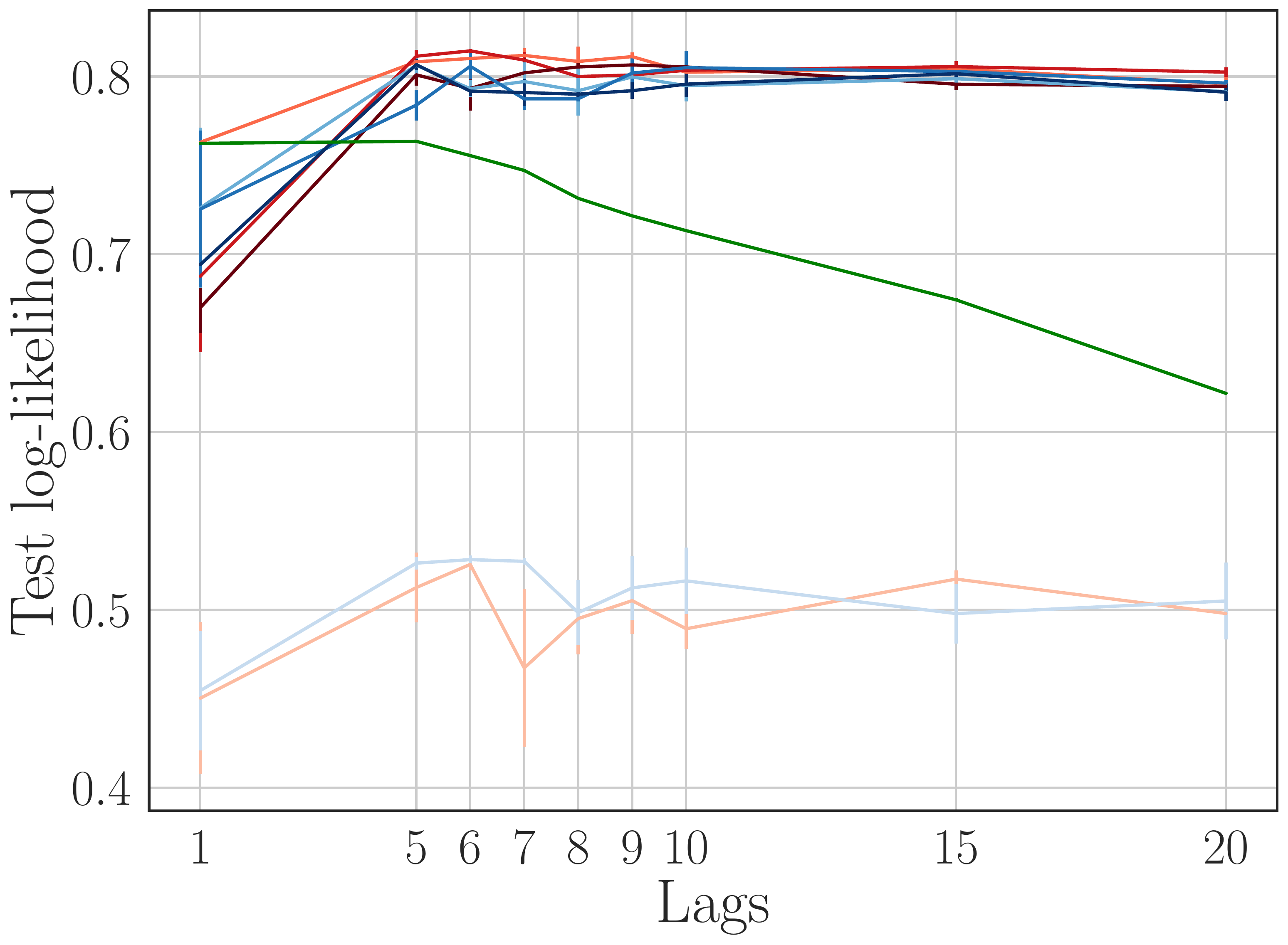}
        \caption{NASCAR.}
        \label{app:fig:syn:nascar_lls}
    \end{subfigure}%
    \hfill%
    \begin{subfigure}[b]{0.48\textwidth}
    	\includegraphics[width=\textwidth]{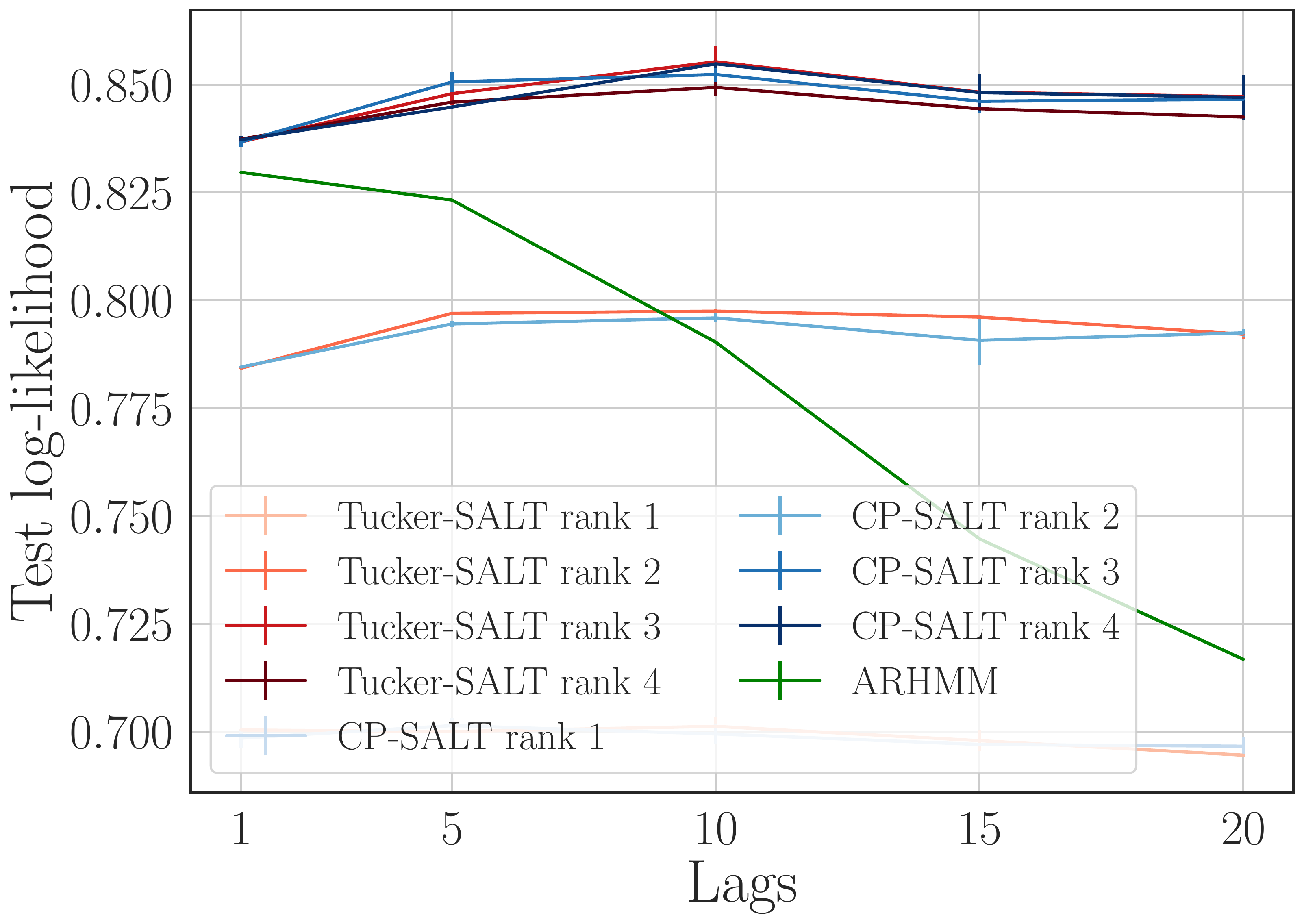}
        \caption{Lorenz.}
        \label{app:fig:syn:lorenz_lls}
    \end{subfigure}
    \caption{Quantitative performance of different \method models and ARHMMs (averaged over 3 different runs) on the synthetic experiments presented in Section \ref{sec:results:syn}. The test-set log likelihood is shown as a function of lags in the \method model, for both \textbf{(A)} the NASCAR\textsuperscript{\textregistered} and \textbf{(B)} Lorenz synthetic datasets.}
    \label{app:fig:syn}
\end{figure}

We include further results and analysis for the NASCAR\textsuperscript{\textregistered} and Lorenz attractor experiments presented in Section \ref{sec:results:syn}.
We compare the marginal likelihood achieved by single-subspace \method models of different sizes.  We see that \method outperforms ARHMMs, and can fit larger models (more lags) without overfitting (Figure \ref{app:fig:syn}).  Note that the SLDS does not admit exact inference, and so we cannot readily compute the exact marginal likelihood for the SLDS.  

\subsection{TVART versus \method in recovering the parameters of SLDSs}
\label{subsec:tvart_comparison}

We compared SALT to TVART \cite{tvart}, another tensor-based method for modeling autoregressive processes.  We modified TVART (as briefly described in the original paper,~\cite{tvart}) so that it can handle AR(p) processes, as opposed to only AR(1) processes.  TVART is also not a probabilistic model (i.e., cannot compute log-likelihoods), and so we focus our comparison on how well these methods recover the parameters of a ground-truth SLDS.

We used the same SLDS that we used to generate the NASCAR\textsuperscript{\textregistered} dataset in Section \ref{sec:results:syn}. We then used $L=7$ CP-\method and Tucker-\method with ranks 3 and 2, respectively, and computed the MSE between the ground truth tensor and \method tensors. For TVART, we used $L=7$, bin size of 10, and ranks 2 and 3 to fit the model to the data. We then clustered the inferred dynamics parameters to assign discrete states. To get the TVART parameter estimation, we computed the mean of the dynamics parameters for each discrete state and computed the MSE against the ground truth tensor. The MSE results are as follows:

\begin{table}[h!]
\centering
\caption{Results comparing \method and TVART~\cite{tvart} on the NASCAR example.}
\label{app:tab:tvart}
\begin{tabular}{@{}lccr@{}}
\toprule
\textbf{Model}                  & \textbf{Rank} & \textbf{Tensor Reconstruction MSE} ($\times 10^{-3}$) & \textbf{Number of parameters} \\ \midrule
TVART                           & 2    & 0.423                                       & 1.4K             \\
TVART                           & 3    & 0.488                                       & 2.0K             \\
Tucker-\method                  & 2    & 0.294                                       & 0.6K                 \\
CP-\method                      & 3    & 0.297                                       & 0.7K                 \\ \bottomrule
\end{tabular}
\end{table}

Table \ref{app:tab:tvart} shows that \method models recover the dynamics parameters of the ground truth SLDS more accurately.  Furthermore, we see that \method models use fewer parameters than TVART models for the dataset (as the number of parameters in TVART scales linearly with the number of windows).  We also note that TVART cannot be applied to held-out data, and, without post-hoc analysis, does not readily have a notion of re-usable dynamics or syllables.

\subsection{The effect of the number of switches on the recovery of the parameters of the autoregressive dynamic tensors}
\begin{figure*}[!t]
\begin{center}
	\includegraphics[width=0.5\textwidth]{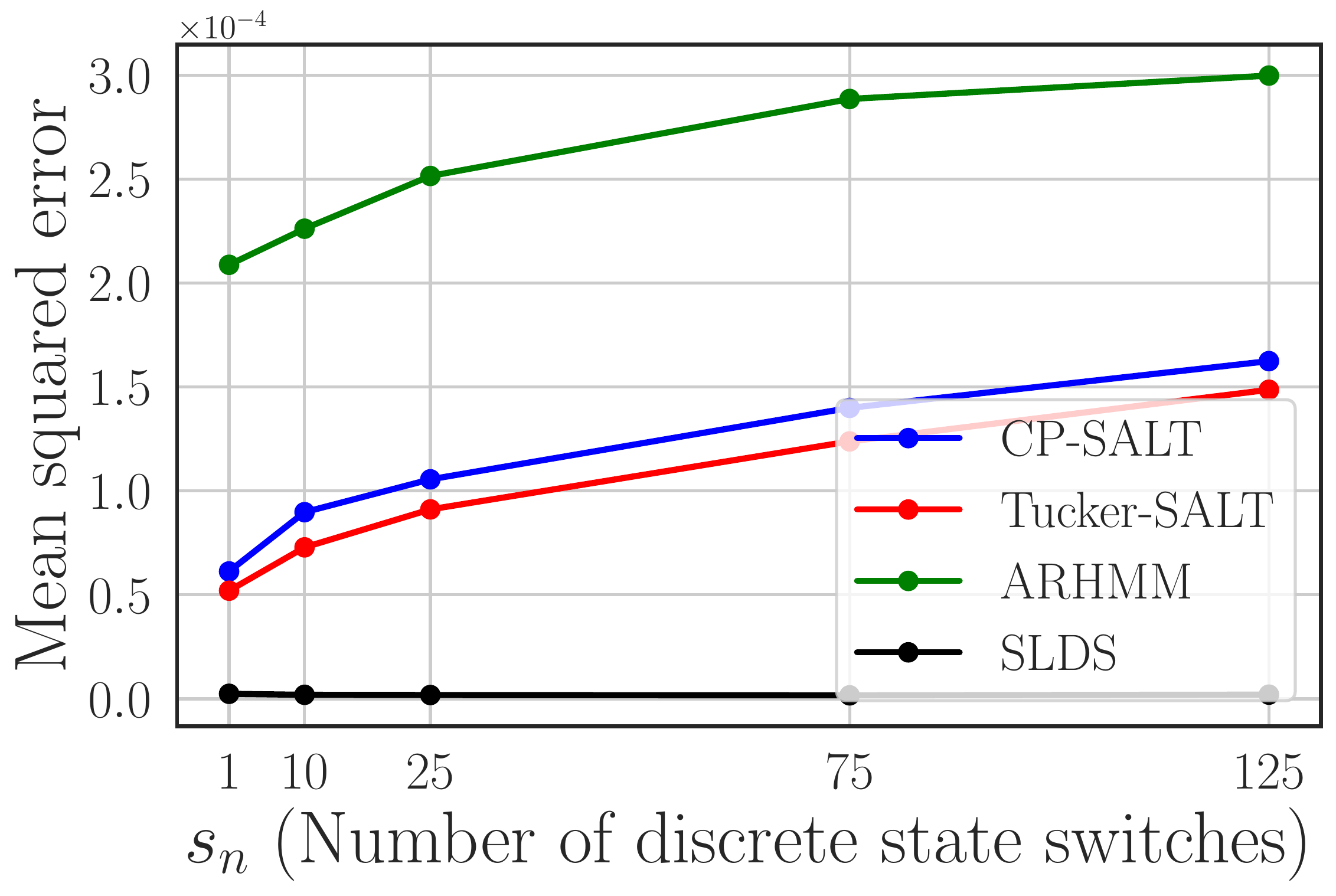}
    \caption{\textbf{The quality of \method approximation of SLDSs decreases as the number of discrete state switches increases:} The data comes from an SLDS with $H=2$, $N=20$, and $D=7$. 15,000 timesteps were generated, with varying numbers of evenly spaced discrete state switches (x-axis). The mean squared error of reconstructing the autoregressive tensors increased as a function of the number of discrete state switches. Note that we combined the 3rd order autoregressive tensors from each discrete state into a 4th order tensor, and calculated the MSE based on these 4th order tensors.}
    \label{figsupp:n_switches}
\end{center}
\end{figure*}

We asked how the frequency of discrete state switches affected SALT's ability to recover the autoregressive tensors.
We trained CP-SALT, Tucker-SALT, the ARHMM, all with $L=5$ lags, and the SLDS on data sampled from an SLDS with varying number of discrete state switches. The ground-truth SLDS model had $H=2$ discrete states, $N=20$ observations and $D=7$ dimensional continuous latent states.
The matrix $\mbA^{(h)}(\mbI - \mbK^{(h)}\mbC^{(h)})$ of each discrete state of the ground-truth SLDS had 1 real eigenvalue and 3 pairs of complex eigenvalues.
We sampled 5 batches of $T=15,000$ timesteps of data from the ground-truth SLDS, with $s_n \in \{1, 10, 25, 75, 125\}$ number of discrete state switches that were evenly spaced out across the data.
We then computed the mean squared error (MSE) between the SLDS tensors and the tensors reconstructed by \method, the ARHMM, and the SLDS. (Figure \ref{figsupp:n_switches}). More precisely, we combined the 3rd order autoregressive tensors from each discrete state into a 4th order tensor, and calculated the MSE based on these 4th order tensors. 
As expected, the MSE increased with the number of switches in the data, indicating that the quality of \method approximation of SLDSs decreases as the frequency of discrete state switches increases.

\clearpage
\newpage

\section{Modeling Mouse Behavior}
\label{app:sec:mouse}
We include further details for the mouse experiments in Section \ref{sec:results:mouse}.

\subsection{Training Details}
We used the first 35,949 timesteps of data from each of the three mice, which were collected at 30Hz resolution. 
We used $H = 50$ discrete states and fitted ARHMMs and CP-\method models with varying lags and ranks.
Similar to \citet{wiltschko2015mapping}, we imposed stickiness on the discrete state transition matrix via a Dirichlet prior with concentration of 1.1 on non-diagonals and $6\times 10^4$ on the diagonals. 
These prior hyperparameters were empirically chosen such that the durations of the inferred discrete states and the given labels were comparable.
We trained each model 5 times with random initialization for each hyperparameter, using 100 iterations of EM on a single NVIDIA Tesla P100 GPU.

\subsection{Video Generation}
Here we describe how the mouse behavioral videos were generated.
We first determined the CP-\method hyperparameters as those which led to the highest log-likelihood on the validation dataset. Then, using that CP-\method model, we computed the most likely discrete states on the train and test data.
Given a discrete state $h$, we extracted slices of the data whose most likely discrete state was $h$.
We padded the data by 30 frames (i.e. 1 second) both at the beginning and the end of each slice for the movie.
A red dot appears on each mouse for the duration of discrete state $h$.
We generated such videos for all 50 discrete states (as long as there existed at least one slice for each discrete state) on the train and test data.
For a given discrete state, the mice in each video behaved very similarly (e.g., the mice in the video for state 18 ``pause" when the red dots appear, and those in the video for state 32 ``walk" forward), suggesting that CP-\method is capable of segmenting the data into useful behavioral syllables.
See ``MoSeq\_salt\_videos\_train" and ``MoSeq\_salt\_videos\_test" in the supplementary material for the videos generated from the train and test data, respectively.
``salt\_crowd\_$i$.mp4" refers to the crowd video for state $i$.
We show the principal components for states $1, 2, 13, 32, 33, 47$ in Figure \ref{fig:moseq_result_supp}.  

\subsection{Modeling Mouse Behavior: Additional Analyses}
\label{app:sec:mouse:additional}
\begin{figure}[!h]
\begin{center}
	\includegraphics[width=\textwidth]{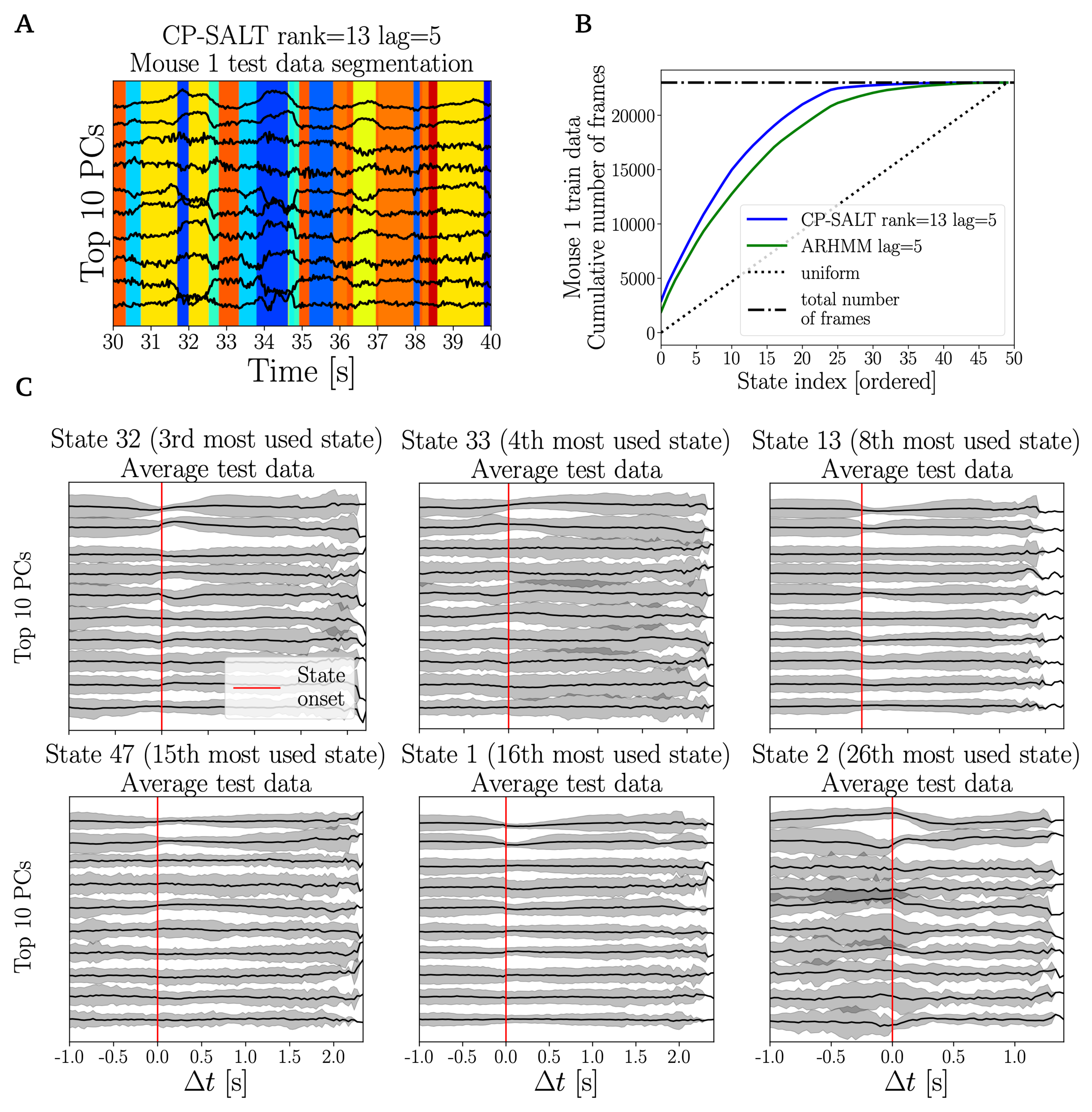}
    \caption{\textbf{CP-\method leads to qualitatively good segmentation of the mouse behavioral data into distinct syllables.}: \textbf{(A)} 30 seconds of test data (Mouse 1) with the discrete states inferred by CP-\method as the background color. \textbf{(B)} For one mouse, the cumulative number of frames that are captured by each discrete state, where the discrete states are ordered according to how frequently they occur. \textbf{(C)} The average test data, with two standard deviations, for six states of CP-\method, aligned to the time of state onset. The shrinkage of the gray region around the state onset tells us that CP-\method segments the test data consistently.}
    \label{fig:moseq_result_supp}
\end{center}
\end{figure}

We also investigated whether \method qualitatively led to a good segmentation of the behavioral data into discrete states, shown in Figure \ref{fig:moseq_result_supp}.
Figure \ref{fig:moseq_result_supp}A shows a 30 second example snippet of the test data from one mouse colored by the discrete states inferred by CP-\method. 
CP-\method used fewer discrete states to model the data than the ARHMM (Figure \ref{fig:moseq_result_supp}B). Coupled with the finding that CP-\method improves test-set likelihoods, this  suggests that the ARHMM may have oversegmented the data and CP-\method may be better able to capture the number of behavioral syllables.
Figure \ref{fig:moseq_result_supp}C shows average test data (with two standard deviations) for a short time window around the onset of a discrete state (we also include mouse videos corresponding to that state in the supplementary materials).
The shrinking gray area around the time of state onset, along with the similar behaviors of the mice in the video, suggests that CP-\method is capable of segmenting the data into consistent behavioral syllables.

\clearpage
\newpage

\section{Modeling \textit{C. elegans} Neural Data}
\label{sec:celegans_appendix}
We include further details and results for the \emph{C. elegans} example presented in Section \ref{sec:results:ce}.  This example highlights how \method can be used to gain scientific insight in to the system. 

\subsection{Training Details}
We used $\sim$3200 timesteps of data (recorded at 3Hz) from one worm, for which 48 neurons were confidently identified.
The data were manually segmented in to seven labels (reverse sustained, slow, forward, ventral turn, dorsal turn, reversal (type 1) and reversal (type 2).
We therefore used $H = 7$ discrete states in all models (apart from the GLM).
After testing multiple lag values, we selected $L=9$ for all models, as these longer lags allow us to examine longer-timescale interactions and produced better segmentations across models, with only a small reduction in variance explained.
We trained each model 5 times with KMeans initialization, using 100 iterations of EM on a single NVIDIA Tesla V100 GPU.
Models that achieved 90\% explained variance on a held-out test set were then selected and analyzed (similar to \citet{linderman2019hierarchical}).

\subsection{Additional Quantitative Results}
\begin{figure}[h!]
\begin{center}
	\includegraphics[width=0.8\textwidth]{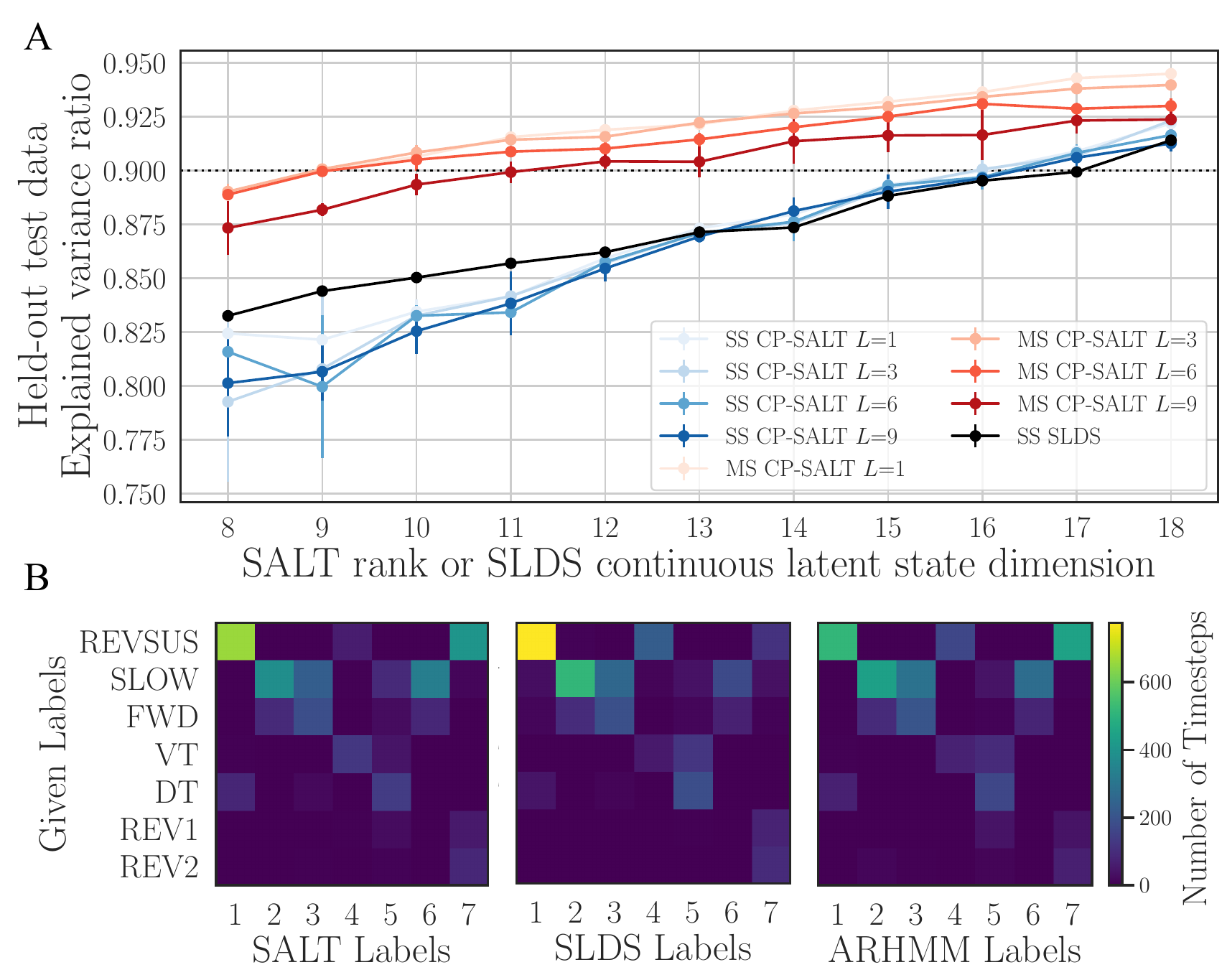}
    \caption{\textbf{\method and SLDS perform comparably on held-out data}: \textbf{(A)}: Explained variance on a held-out sequence.  Single-subspace (SS) \method and SLDS perform comparably.  Multi-subspace (MS) \method achieves a higher explained variance with fewer ranks.  Multi-subspace SLDS was numerically unstable.  \textbf{(B)}: Confusion matrices between given labels and predicted labels.  All methods produce similar quality segmentations. }
    \label{figsupp:celeagns_supp}
\end{center}
\end{figure}

Figure \ref{figsupp:celeagns_supp} shows additional results for training different models.  In Figure \ref{figsupp:celeagns_supp}A we see that models with larger ranks (or latent dimension) achieve higher explained variance.  Interestingly, longer lags can lead to a slight reduction in the explained variance, likely due to overfitting.  This effect is less pronounced in the more constrained single-subspace \method, but, these models achieve lower explained variance ratios throughout.  Longer lag models allow us to inspect longer-timescale dependencies, and so are more experimentally insightful.  Figure \ref{figsupp:celeagns_supp}B shows the confusion matrix for discrete states between learned models and the given labels.  The segmentations were similar across all models that achieved 90\% explained variance.

\subsection{Additional Autoregressive Filters}
\begin{figure}[!h]
\begin{center}
	\includegraphics[width=0.95\textwidth]{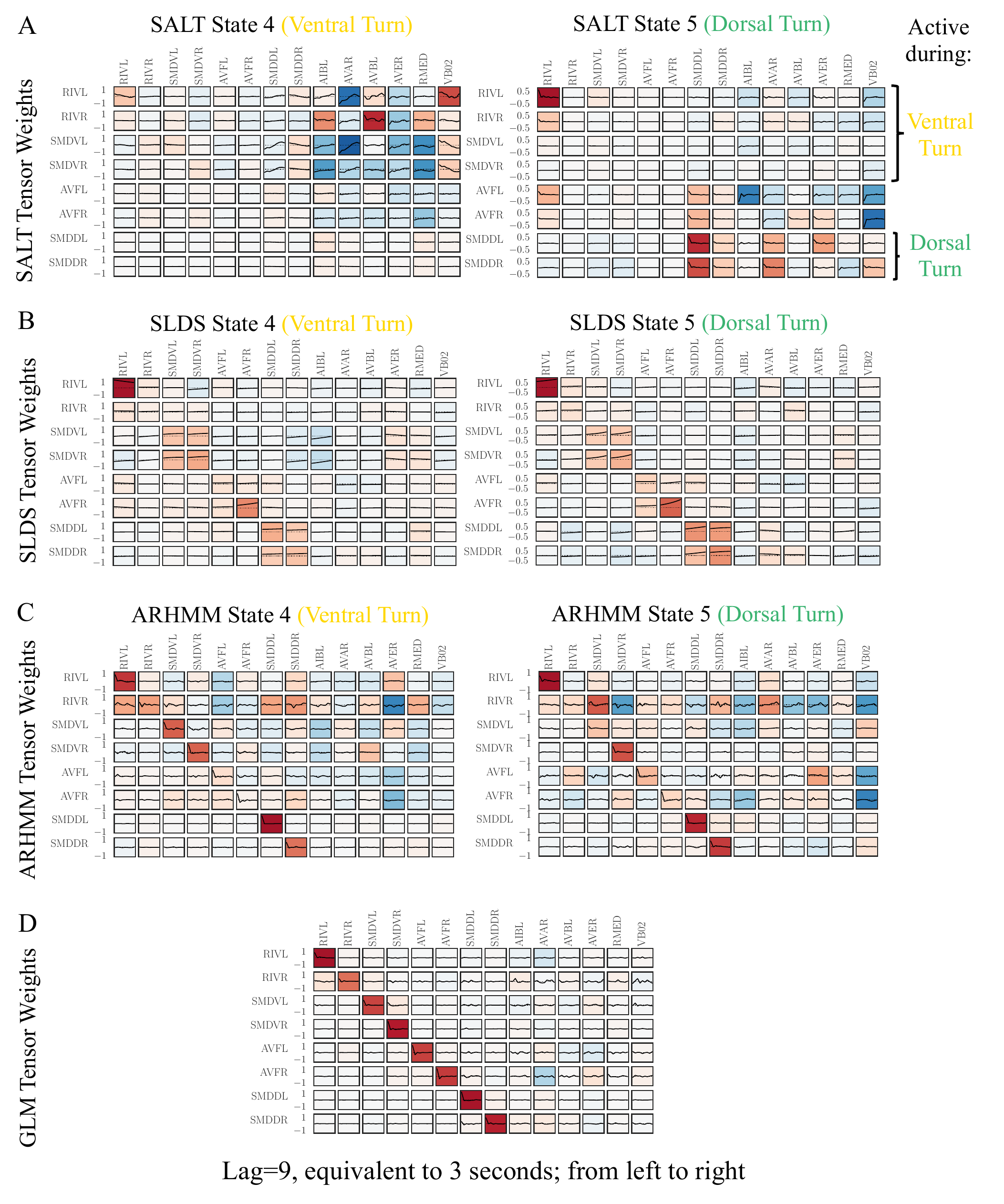}
    \caption{\textbf{Autoregressive tensors learned by different models (Ventral and Dorsal Turns)}: \textbf{(A-C)} One-dimensional autoregressive filters learned in two states by \method, SLDS, ARHMM (identified as ventral and dorsal turns), and \textbf{(D)} by a GLM. \texttt{RIV} and \texttt{SMDV} are known to mediate ventral turns, while \texttt{SMDD} mediate dorsal turns~\citep{kato2015global, gray2005circuit, yeon2018sensory}.  }
    \label{figsupp:celeagns_supp-1}
\end{center}
\end{figure}

\begin{figure}[!h]
\begin{center}
	\includegraphics[width=0.95\textwidth]{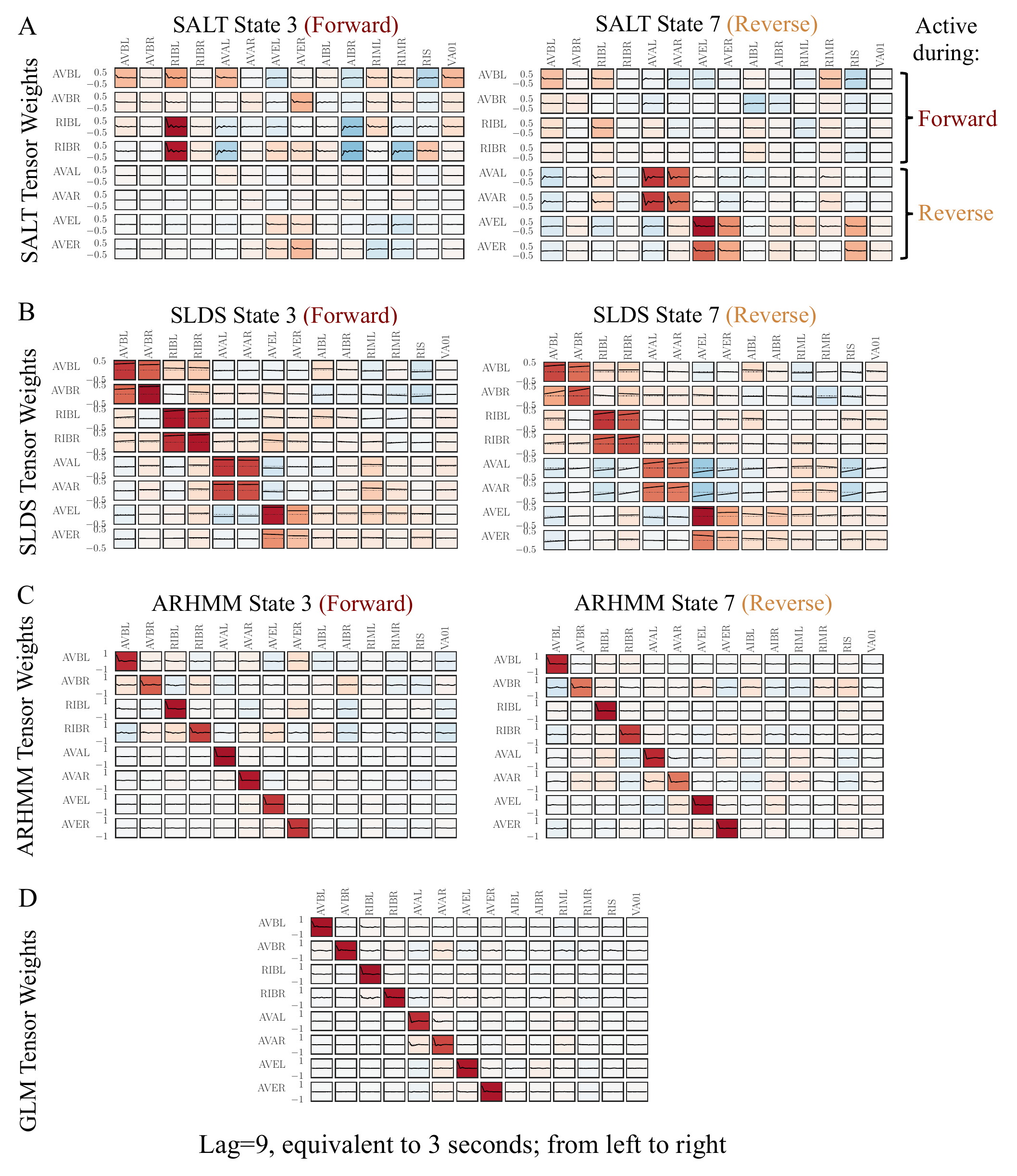}
    \caption{\textbf{Autoregressive tensors learned by different models (Forward Locomotion and Reversal)}: \textbf{(A-C)} One-dimensional autoregressive filters learned in two states by \method, SLDS, ARHMM (identified as forward and reverse), and \textbf{(D)} by a GLM. \texttt{AVB} and \texttt{RIB} are known to mediate forward locomotion, while \texttt{AVA} and \texttt{AVE} are involved in initiating reversals~\citep{kato2015global, gray2005circuit, chalfie1985neural, piggott2011neural}. }
    \label{figsupp:celeagns_supp-2}
\end{center}
\end{figure}

Figures \ref{figsupp:celeagns_supp-1} and \ref{figsupp:celeagns_supp-2} show extended versions of the autoregressive filters included in Section \ref{sec:results:ce}.  Figure \ref{figsupp:celeagns_supp-1} shows the filters learned for ventral and dorsal turns (for which panel A was included in Figure \ref{fig:celegans_analysis}), while Figure \ref{figsupp:celeagns_supp-2} shows the filters for forward and backward locomotion.  Note that the GLM does not have multiple discrete states, and hence the same filters are used across states.  We see for ARHMM and \method that known-behavior tuned neurons have higher magnitude filters (determined by area under curve), whereas the SLDS and GLM do not recover such strong state-specific tuning. Since the learned SLDS did not have stable within-state dynamics, the autoregressive filters could not be computed using \Cref{app:equ:truncated_kalman_form}. We thus show $C A^{(h) l} C^+$ for lag $l$, where $C^+$ denotes the Moore-Penrose pseudoinverse of $C$, as a proxy for the autoregressive filters of discrete state $h$ of the SLDS. Note that this is a post-hoc method and does not capture the true dependencies in the observation space.  

We see that \method consistently assigns high autoregressive weight to neurons known to be involved in certain behaviors (see Figures \ref{figsupp:celeagns_supp-1} and \ref{figsupp:celeagns_supp-2}).  In contrast, the ARHMM identifies these relationships less reliably, and the estimate of the SLDS autoregressive filters identifies few strong relationships.  As the GLM only have one ``state'', the autoregressive filters are averaged across state, and so few strong relationships are found.  This highlights how the low-rank and switching properties of \method can be leveraged to glean insight into the system.

\end{document}